\documentclass{article}
\ifdefined\pdfsuppressptexinfo \pdfsuppressptexinfo=-1 \fi
\ifdefined\pdfinfoomitdate \pdfinfoomitdate=1 \fi
\ifdefined\pdftrailerid \pdftrailerid{} \fi

\usepackage[T1]{fontenc}
\usepackage{microtype}
\usepackage{graphicx}
\usepackage{booktabs,tabularx,array}
\usepackage{amsmath,amsthm,mathtools}
\usepackage{xcolor}
\usepackage{enumitem}
\usepackage{algorithm}
\usepackage{algpseudocode}
\usepackage{capt-of}
\usepackage{tikz}
\usetikzlibrary{arrows.meta,positioning,calc,fit,patterns,decorations.pathreplacing}
\usepackage[hidelinks]{hyperref}

\usepackage[preprint]{icmlfallback}
\IfFileExists{newtxmath.sty}{\usepackage{newtxmath}}{\usepackage{txfonts}}

\setlist[itemize]{leftmargin=1.1em,itemsep=1pt,topsep=2pt,parsep=0pt}
\setlist[enumerate]{leftmargin=1.4em,itemsep=1pt,topsep=2pt,parsep=0pt}

\theoremstyle{definition}
\newtheorem{assumption}{Assumption}
\theoremstyle{plain}
\newtheorem{lemma}[assumption]{Lemma}
\newtheorem{theorem}[assumption]{Theorem}
\newtheorem{corollary}[assumption]{Corollary}
\newtheorem{proposition}[assumption]{Proposition}

\newcommand{\E}{\mathbb{E}}
\newcommand{\KL}{\mathrm{KL}}

\newcommand{\cG}{\mathcal{G}}
\newcommand{\one}{\mathbf{1}}
\newcommand{\dec}{\mathrm{dec}}
\newcommand{\stor}{\mathrm{stor}}
\newcommand{\KV}{\mathrm{KV}}
\newcommand{\range}{\mathrm{range}}
\newcommand{\cA}{\mathcal{A}}

\icmltitlerunning{When Thoughts Become Memory: Unifying In-Context Learning and Fine-Tuning}

\begin{document}

\twocolumn[
\icmltitle{When Thoughts Become Memory:\\Unifying In-Context Learning and Fine-Tuning}
\icmlsetsymbol{equal}{*}
\begin{icmlauthorlist}
\icmlauthor{Leon Chlon}{hl}
\icmlauthor{Julie Huang}{hl}
\end{icmlauthorlist}
\icmlaffiliation{hl}{Hassana Labs}
\icmlkeywords{in-context learning, fine-tuning, information theory, code length, held-out validation}
\vskip 0.3in
]
\printAffiliationsAndNotice{}

\begin{abstract}
We are the first to give a certified rule for whether unseen data belong in a pretrained model's context or its weights, built on an exact law for their information value. That value, the code length saved on held-out queries, equals in expectation the bits stored about the query, plus a prior repair, minus a decoding loss, for any model, architecture, dataset and task. The law's ceilings certify the choice from one measurement, and an anytime-valid test makes it from held-out queries, with error controlled whenever it stops. Every certificate the ceilings issue on a trained transformer, TabPFN, an occupancy network and a stereo network is correct. The test picks TabPFN over XGBoost and a trained surrogate over an in-context learner on airfoil pressure, calls the better of GPT-2's cache and fine-tune in 61,948 of 64,000 query streams over 32 comparisons, and errs in at most 3.3\% of null streams. The law also diagnoses when learning in context helps and when it does not. Deleting part of a context is held-out validation of that part in any model, and adding parts one at a time traces the learning curve in context: one patch that sees a shape feature cuts an occupancy network's code length there a hundredfold, while eight that miss it nearly double it, and one camera that sees the surface beats four in a stereo network. Ranked by this label-free valuation, a kriging learner's three best support cars beat four drawn at random, and on a yes-or-no claim the valuation recovers a published evidence adjudicator, with AUROC 0.91.
\end{abstract}

\begin{table*}[t]
\caption{The results at a glance: what each tool takes, what it guarantees, and how the paper tests and checks it.}\label{tab:summary}
\centering\footnotesize
\setlength{\tabcolsep}{4pt}
\renewcommand{\arraystretch}{1.1}
\begin{tabularx}{\textwidth}{@{}>{\raggedright\arraybackslash}p{0.14\textwidth}>{\raggedright\arraybackslash}p{0.17\textwidth}>{\raggedright\arraybackslash}X>{\raggedright\arraybackslash}p{0.35\textwidth}@{}}
\toprule
Tool & What it takes & What it guarantees & Tested on, and checked by\\
\midrule
\multicolumn{4}{@{}l}{\emph{With the information terms of the data}}\\
Exact law (Thm.~\ref{thm:law}) & held-out code lengths & $\E[\Gamma_c]=I(Q;Z_c)+\KL_0-\dec_c$, in context and in the weights alike & every term in closed form for a linear-Gaussian model, matched by simulation (App.~\ref{sec:closedforms})\\
Certificates (Prop.~\ref{prop:choose}(ii)) & one channel's code length, the other's ceiling & the channel beats every memory confined to the window, or of at most $C$ bits & 25 of 64 on a trained transformer and 36 of 96 on TabPFN, occupancy and stereo networks, all correct\\
\midrule
\multicolumn{4}{@{}l}{\emph{From the model's probabilities and reruns alone}}\\
Decoding-loss bound (Prop.~\ref{prop:choose}(iii)) & a fine-tune's advantage, context in the window & no fine-tune adds more than the decoding loss in context & GPT-2: at least 0.35--0.55 bits per token, prior repair the cache leaves unrealised\\
Anytime-valid test (Thm.~\ref{thm:eprocess}) & held-out code lengths of both channels & a wrong call with probability at most $\alpha$, whenever the test stops & tables, airfoil pressure and GPT-2; at most 3.3\% wrong calls on nulls, within 5\%; a power curve; the same call in 200 reorderings\\
Prequential curves (Prop.~\ref{prop:preq}) & the code lengths of the context's own units & each equals the held-out learning curve at that size, in expectation & GPT-2: within 1.97 standard errors of the held-out curves; sums within 0.1\% and 1.5\%\\
Storage budget (Thm.~\ref{thm:storage}) & in-context outputs over resampled contexts & a fixed state within $\varepsilon$ bits of the context memory stores at least $I(\hat Y;D)-\varepsilon$ bits & GPT-2 on tables of facts: the estimate stays below the exact 5.14 bits over all 4,096 tables\\
Value of a part of the context (Cor.~\ref{cor:ablation}) & a rerun without the part & its exact leave-one-out value, label-free as $I_G$, split into content and form & occupancy, stereo and kriging models; claims fixed in advance, matched controls, native stereo and kriging code reproduced exactly in 12 and 168 comparisons\\
Evidence budget (Cor.~\ref{cor:sharp}) & a claim's probability with and without its evidence & the bits a claim needs, against the bits its evidence supplies \citep{chlon2026predictable} & two language-model verifiers on VitaminC claims, AUROC 0.91\\
\bottomrule
\end{tabularx}
\end{table*}

\section{Introduction}\label{sec:intro}
A pretrained model holds unseen data in one of two places. Read as context, they live in a state the frozen model computes afresh in every forward pass: the feature grid of an occupancy network, the cost volume of a stereo network, the kriging weights of a kriging learner, the key--value cache of a language model. Written by fine-tuning, they become part of the weights. Every model that reads context poses the same question: when should a thought held in context become a memory in the weights? Current accounts of in-context learning, as implicit Bayesian inference \citep{xie2022explanation} or as gradient descent carried out by attention \citep{vonoswald2023,akyurek2023learning}, describe stylised models of language and of linear regression, and give no quantity to answer it with.

We resolve it with an information-theoretic construction. Context and weights are both valued by the bits they save the model on held-out queries, and both obey one exact law that separates what a memory stores from what its decoding loses. The two channels can then be compared term by term, and capacity bounds on each term become certificates that choose a channel from one measurement. Deleting part of the context and rerunning the model gives the value of that part in any model, and with it a test of whether evidence supports a claim. The construction asks a model for two things only: a probability for each outcome and a rerun on part of its context. Every tool runs on these two, and the certificates also need exact information terms. Nothing else is needed from the dataset, the architecture or the task, so a hosted model supplies everything through its log-probabilities, and a local model adds its weights, which give the certificates their capacity bounds. We prove what follows from this construction and validate it on language, vision, fluid-dynamics and tabular models.

In expectation the saving splits exactly into three terms, for the context memory and the weights alike (Theorem~\ref{thm:law}): the information the memory stores about the query; a prior repair, which corrects the model's prior over outputs and is available to any context, even one that says nothing about the query; and a decoding loss, by which the model's prediction falls short of the best prediction from its memory. The split refines the resolution--reliability decomposition of the logarithmic score \citep{brocker2009reliability} by conditioning on the memory a channel builds. That conditioning makes the two channels comparable: it separates what a memory stores from what its decoder uses, and when the context fits the window it caps what any fine-tune can add at the decoding loss in context. No previous work has made context and weights comparable this way or turned the comparison into a certified rule.

The law decides between the two. Beyond the prior repair they share, a fine-tune cannot save more bits than it stores, and a context memory no more than its window holds, so one channel's measured gain, set against the other's ceiling, can certify the choice. A full fine-tune opens no new space for what it learns: it writes into the bits that already hold the model. Prequential learning curves, read from the context itself, and an anytime-valid test on held-out queries make the choice against any model trained on the same data, with error controlled whenever the test stops. A storage budget sets how many bits fixed weights must store to reproduce the predictions made in context (Section~\ref{sec:choose}).

The choice follows the data, to context or to the weights. On two OpenML tables the test picks TabPFN, which learns from the rows in its context, over gradient-boosted trees trained on the same 128 rows, in every seed. On airfoil pressure fields it picks a trained surrogate over an in-context learner in every seed, and with 32 designs instead of 128 it picks the in-context learner in two seeds of three. On simulated streams with no true winner it calls one in at most 3.3\% of them, within its 5\% limit. Where the information terms are exact, the ceilings issue 25 of 64 possible certificates on a trained transformer and 36 of 96 on TabPFN, an occupancy network and a stereo network, and every one agrees with the measured winner (Section~\ref{sec:evidence}).

The law also tells when learning in context helps. Deleting a part of the context and rebuilding the memory from the rest is held-out validation of that part, its exact leave-one-out value, run in the forward pass with the weights untouched in any architecture, and a matched control of the same form separates what the part says from the form it takes.

Measured this way, the bits follow the evidence each query depends on. In an occupancy network, the 300 points that observe a local shape feature are worth 17.66 bits per query, while 300 matched points from unchanged geometry are worth 0.001 bits. In a stereo network, a camera that sees the queried surface carries three times the bits of one that does not. The learning curve in context turns on which items arrive. One patch that observes the feature takes the occupancy network's code length at the feature from 4.99 to 0.05 bits per site, while eight patches that miss the feature raise it to 9.06, so more context makes the prediction worse. A kriging learner that ranks its support cars by its own valuation codes better with three than with four drawn at random. For two language-model verifiers, the law recovers the Trace Budget adjudicator \citep{chlon2026predictable}: the bits a binary claim needs, set against the bits its evidence supplies, separate supporting from insufficient evidence with AUROC 0.91 (Section~\ref{sec:valuing}).

\paragraph{Contributions.} Table~\ref{tab:summary} states the results.
\begin{itemize}
\item Compare context and weights term by term, through an exact law for the expected gain of a context memory and of a fine-tune (Section~\ref{sec:law}).
\item Certify the choice against a whole class from one measured code length, every memory confined to a window or every state of at most $C$ bits, without training it; and, when the context fits the window, bound a model's decoding loss from below with one fine-tune (Section~\ref{sec:choose}).
\item Value each part of a context exactly, and without labels, in any model, separating what it says from its form and tracing learning curves in context (Section~\ref{sec:valuing}).
\item Choose with error controlled at any stopping time when the information terms are unknown, with evidence across language, vision, fluid-dynamics and tabular models (Section~\ref{sec:evidence}). The appendix writes a transformer's cache exactly as a weight write with query-dependent step sizes and gives a linear-Gaussian model in which every term is closed form.
\end{itemize}

\section{Learning from context}\label{sec:setup}
All logarithms are base two and all information is in bits, except in Appendices~\ref{sec:token}--\ref{sec:cache} and their proofs, which work with logits in nats. The code length of a query $Q$ given a context $c$ is $\ell_\theta(Q\mid c)=-\log_2p_\theta(Q\mid c)$, the bits the model needs to encode the observed outcome: for occupancy at a site, a depth at a pixel or a pressure at a surface point, the negative log-probability or log-density of the observed value, and for a sequence $Q=(q_1,\dots,q_m)$, the sum $-\sum_{t}\log_2p_\theta(q_t\mid c,q_{<t})$.

A frozen model $\theta_0$ with base predictive $p_0=p_{\theta_0}(\cdot\mid\varnothing)$ can absorb a dataset $D$ in two ways. \textbf{In context} (channel A), it computes from $D$ a memory $Z_A$, such as a feature grid, a cost volume, a set of kriging weights or a key--value cache, and predicts $p_A(Q\mid D)=p_{\theta_0}(Q\mid Z_A)$. \textbf{By fine-tuning} (channel B), it writes $D$ into its weights, $Z_B=\theta_T(D)$, and predicts $p_B(Q\mid D)=p_{\theta_T(D)}(Q\mid\varnothing)$. A learner without trained weights, such as a kriging learner, has only channel A. Each channel's gain is
\begin{equation}\label{eq:gain}
\Gamma_c(Q;D)=\ell_{\theta_0}(Q\mid\varnothing)-\ell_c(Q\mid D)=\log_2\frac{p_c(Q\mid D)}{p_0(Q)}.
\end{equation}
For a transformer the context memory is the key--value cache (the cache, from here on), and two teacher-forced passes, one over $Q$ and one over $D$ followed by $Q$, measure its gain, $\Gamma_A=\ell(D)+\ell(Q)-\ell(D\oplus Q)$. For a part $G$ of the context and the rest $R$, $D=(R,G)$, deleting $G$ and rebuilding the memory from $R$ gives the ablation gain of $G$,
\begin{equation}\label{eq:ablation}
\Delta_G(Q)=\ell_c(Q\mid R)-\ell_c(Q\mid D)=\Gamma_c(Q;D)-\Gamma_c(Q;R).
\end{equation}

\begin{assumption}[Clean memory]\label{ass:clean}
Each memory is built from its own data and exogenous randomness, so that $Q\perp Z_c\mid D$. After a deletion the memory is rebuilt from $R$ alone, and no state computed from $G$ is reused.
\end{assumption}

A deletion reruns the computation that builds the memory on what remains: the encoder of the feature grid, the construction of the cost volume, the fit of the kriging learner, and in a transformer every cache entry after the deleted span, because those entries have already absorbed the span, and masking attention to it would leave its information in place. The rebuilt memory carries no trace of the deleted evidence (Lemma~\ref{lem:exclusion} in Appendix~\ref{app:lemma}), so a deletion is valued at what the evidence adds beyond the rest, as held-out validation values a training example at what the rest of the training set does not supply.

\section{The law of in-context gain}\label{sec:law}
The truth is $(D,Q)\sim P^*$, with true posterior predictive $P^*(Q\mid D)$ and marginal $P^*_Q$.

\begin{theorem}[Exact law]\label{thm:law}
Under Assumption~\ref{ass:clean}, for either channel,
\begin{equation}\label{eq:law}
\E[\Gamma_c]=\underbrace{I(Q;Z_c)}_{\text{bits stored}}+\underbrace{\KL(P^*_Q\,\|\,p_0)}_{\text{prior repair}}-\underbrace{\E\,\KL\big(P^*_{Q\mid Z_c}\,\big\|\,p_c(\cdot\mid Z_c)\big)}_{\text{decoding loss }\dec_c}.
\end{equation}
\end{theorem}

\begin{proof}
Pointwise, with $Z=Z_c$,
\[
\log\frac{p_c(Q\mid Z)}{p_0(Q)}=\log\frac{P^*(Q\mid Z)}{P^*(Q)}+\log\frac{P^*(Q)}{p_0(Q)}-\log\frac{P^*(Q\mid Z)}{p_c(Q\mid Z)}.
\]
Take expectations under $P^*$.
\end{proof}

Conditioning on the forecast $F_c=p_c(\cdot\mid Z_c)$ in place of the memory turns the first and last terms into the resolution and reliability of the logarithmic score \citep{brocker2009reliability,weijs2010kullback}, and the decoding loss exceeds that reliability by $I(Q;Z_c\mid F_c)$, the information the memory holds and its decoder leaves unused (Appendix~\ref{app:law}). The ceilings in Section~\ref{sec:choose} act on the bits stored, and the decoding-loss bound on the bits used.

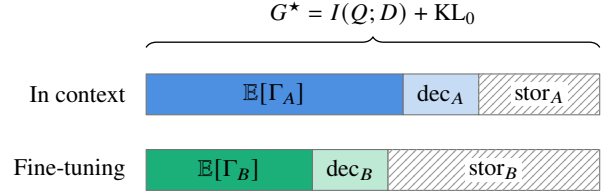
\begin{figure}[t]
\centering
\definecolor{fbBlue}{HTML}{4D8FE0}\definecolor{fbBlueL}{HTML}{C6DCF5}%
\definecolor{fbGreen}{HTML}{1BAF7A}\definecolor{fbGreenL}{HTML}{BDE9D5}%
\begin{tikzpicture}[x=1cm, y=1cm, font=\small, line join=round]
\def\H{0.27}
\tikzset{seg/.style={draw=black!55, line width=0.4pt},
         stor/.style={seg, fill=white, postaction={pattern=north east lines, pattern color=black!40}}}
\draw[decorate, decoration={brace, amplitude=4pt}, line width=0.5pt] (0,0.62) -- (6.0,0.62)
  node[midway, above=5pt] {$G^\star=I(Q;D)+\KL_0$};
\node[anchor=east] at (-0.15,0) {In context};
\draw[seg, fill=fbBlue] (0,-\H) rectangle (3.4,\H);
\draw[seg, fill=fbBlueL] (3.4,-\H) rectangle (4.4,\H);
\draw[stor] (4.4,-\H) rectangle (6.0,\H);
\node at (1.7,0) {$\E[\Gamma_A]$};
\node at (3.9,0) {$\dec_A$};
\node[fill=white, inner sep=1.2pt] at (5.2,0) {$\stor_A$};
\node[anchor=east] at (-0.15,-1.0) {Fine-tuning};
\draw[seg, fill=fbGreen] (0,-1.0-\H) rectangle (2.2,-1.0+\H);
\draw[seg, fill=fbGreenL] (2.2,-1.0-\H) rectangle (3.2,-1.0+\H);
\draw[stor] (3.2,-1.0-\H) rectangle (6.0,-1.0+\H);
\node at (1.1,-1.0) {$\E[\Gamma_B]$};
\node at (2.7,-1.0) {$\dec_B$};
\node[fill=white, inner sep=1.2pt] at (4.6,-1.0) {$\stor_B$};
\end{tikzpicture}
\caption{Both channels, measured in the same bits. The data can give at most $G^\star=I(Q;D)+\KL_0$ bits per query in expectation. Each channel keeps its expected gain and loses the rest in two parts: the storage loss, which its memory drops, and the decoding loss, by which its prediction falls short of the best prediction from its memory (Equation~\eqref{eq:law2}).}
\label{fig:budget}
\end{figure}

Because $Q-D-Z_c$ is a Markov chain, $I(Q;Z_c)=I(Q;D)-I(Q;D\mid Z_c)$, and the law reads
\begin{equation}\label{eq:law2}
\begin{gathered}
\E[\Gamma_c]=G^\star-\underbrace{I(Q;D\mid Z_c)}_{\text{storage loss }\stor_c}-\dec_c,\\
G^\star=I(Q;D)+\KL(P^*_Q\,\|\,p_0)=\E_D\,\KL(P^*_{Q\mid D}\,\|\,p_0).
\end{gathered}
\end{equation}
The storage loss is the information about the query that the memory drops, and $G^\star$ is everything the data can give. Figure~\ref{fig:budget} draws both channels in these terms.

\begin{figure*}[t]
\centering
\includegraphics[width=\textwidth]{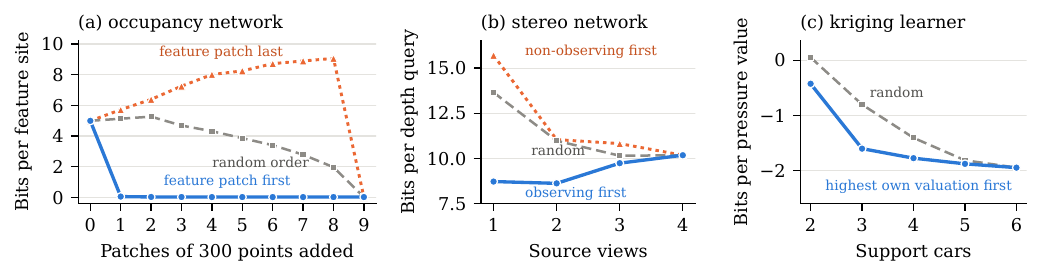}
\caption{Learning curves in context: held-out code length against the number of context items, by order of arrival, with the memory rebuilt from scratch for each set of items. (a) Occupancy network on eight constructed chair pairs: 300 anchor points and up to nine patches of 300 points, one of which observes the feature; code length at the feature sites. (b) MVSNet on five DTU scenes: one to four source views, at the 3,856 pixels that some views observe and others do not, with the views ordered separately at each pixel. (c) Kriging learner on eight DrivAerNet cars: two to six support cars, ranked by the gain $I_G$ its own predictions assign each car; log-density code length, which can be negative. Random orders average four orders, or in (b) all subsets of each size; means over cases.}
\label{fig:counts}
\end{figure*}

\begin{corollary}[Ceiling, channels and content-free context]\label{cor:law}
(i) $\E[\Gamma_c]\le G^\star$, with equality if and only if $p_c(\cdot\mid D)=P^*(\cdot\mid D)$. (ii) For the two channels,
\begin{equation}\label{eq:channels}
\E[\Gamma_A-\Gamma_B]=\E\big[\KL(P^*_{Q\mid D}\,\|\,p_B)-\KL(P^*_{Q\mid D}\,\|\,p_A)\big].
\end{equation}
(iii) If a context $\tilde D$ carries no information about the query, $I(Q;\tilde D)=0$, then $\E[\Gamma_c(Q;\tilde D)]\le\KL(P^*_Q\,\|\,p_0)$.
\end{corollary}

Whichever channel lands closer to the true posterior predictive wins, by exactly the difference of their divergences from it (part (ii)). Part (iii) says what a context with no information about the query can do: repair the prior, and nothing more. This is why demonstrations with random labels still help \citep{min2022rethinking}: they fix the format, the label space and the domain, and that alone is worth up to $\KL(P^*_Q\|p_0)$ bits.

\begin{corollary}[Ablation]\label{cor:ablation}
For $D=(R,G)$, write $\stor_X=I(Q;X\mid Z_X)$ and $\dec_X$ for the storage and decoding losses of the memory built from $X$. Then
\begin{equation}\label{eq:ablationlaw}
\E[\Delta_G]=I(Q;G\mid R)+\big(\stor_R-\stor_D\big)+\big(\dec_R-\dec_D\big).
\end{equation}
The prior repair cancels. The first term is the information $G$ carries about the query beyond the rest, and the storage terms vanish when the memory keeps its context, as a transformer's cache does within its window (Proposition~\ref{prop:capacity}).
\end{corollary}

A matched deletion $\tilde G$ removes a part with the form of a relevant one and content that says nothing about the query beyond the rest, $I(Q;\tilde G\mid D\setminus\tilde G)=0$. Its expected gain, $\E[\Delta_{\tilde G}]=(\stor_{D\setminus\tilde G}-\stor_D)+(\dec_{D\setminus\tilde G}-\dec_D)$, is the cost of the form alone. A relevant deletion of the same form differs from it only in content, so content accounts for the whole difference between the two gains, whichever terms of Equation~\eqref{eq:ablationlaw} carry it: removing what the decoder reads changes its decoding as well as its information. With Corollary~\ref{cor:law}(iii), the pair tests whether a part of the context helps through what it says or only through its form, including any repair of the format.

\paragraph{The model's own valuation.}
Draw the outcome from the model's own full-context prediction, $Q\sim p_c(\cdot\mid D)$. The full context then has no decoding loss, and the expected ablation gain is
\begin{equation}\label{eq:info}
I_G:=\E_{Q\sim p_c(\cdot\mid D)}\big[\Delta_G(Q)\big]=\KL\big(p_c(\cdot\mid D)\,\big\|\,p_c(\cdot\mid R)\big)\ge 0,
\end{equation}
the code length the model saves on its own full-context continuation. It needs no labels, and it is defined for each context separately. For a fixed context, Corollary~\ref{cor:ablation} has no information or storage term: $I_G$ is the decoding loss of the memory rebuilt from $R$, measured against the full-context prediction. A single deletion can therefore be worth many bits even for a binary outcome, when it overturns a confident prediction.

\section{Valuing context in any memory}\label{sec:valuing}

The law turns context ablation into a measurement for any model that conditions on context. The evaluator (Algorithm~\ref{alg:eval} in Appendix~\ref{sec:design}) asks three things of a model: a predictive distribution or a claim probability, a deletion after which the memory can be rebuilt from what remains (Assumption~\ref{ass:clean}), and the same outputs scored with and without the deleted part. Nothing in it is specific to an architecture. For each deletion it records $I_G$ and, for each claim fixed in advance, the claim's probability with and without the deleted evidence, and it pairs each relevant deletion with a matched control of the same form (Section~\ref{sec:law}). We run it on four memories (Table~\ref{tab:contract} in Appendix~\ref{sec:results}): the feature grid of a pretrained convolutional occupancy network, ConvONet \citep{peng2020convonet}, on constructed chairs and ShapeNet objects \citep{chang2015shapenet}; the cost volume of a multi-view stereo network, MVSNet \citep{yao2018mvsnet}, on DTU scenes \citep{jensen2014dtu}; the kriging weights of a kriging learner on DrivAerNet cars \citep{elrefaie2024drivaernet}; and the key--value caches of two language-model verifiers on VitaminC \citep{schuster2021vitaminc}.

Deleting the 300 points that observe a local shape feature and rebuilding the grid drops the mean occupancy probability at the feature sites from 0.9827 to 0.0014 and costs 17.66 bits per query, while deleting 300 matched points from unchanged geometry costs 0.001 bits (Figure~\ref{fig:evidence}a in Appendix~\ref{sec:results}). Deleting a source view that observes the queried surface point costs 1.62 bits per depth query, against 0.55 bits for one that does not, and takes 9.52 percentage points more off the probability that the depth lies within five depth units of its full-context estimate, with the larger loss in all five scenes (Figure~\ref{fig:evidence}b).

Adding a context's items one at a time traces the model's learning curve in context. Each step is the ablation gain of the item added. The information an item adds is never negative, so by Corollary~\ref{cor:ablation} a step that raises the expected loss is one in which the item raised the storage and decoding losses, together, by more than the information it brings.

The curves turn on which items arrive (Figure~\ref{fig:counts}). With 300 anchor points the occupancy network codes the feature sites in 4.99 bits per site. Added first, the one patch that observes the feature takes this to 0.046 bits, while the eight that miss the feature raise it to 9.06, and the feature patch, added after them, brings it down to 0.015. One source view that observes the queried point codes its depth in 8.71 bits, against 13.65 for a view drawn at random and 10.17 for all four views together. The kriging learner can rank its support cars without labels, by the gain $I_G$ its own full-context predictions assign each car. With the top three it codes the pressure better than with four drawn at random, $-1.61$ against $-1.41$ bits per value (a log-density code length can be negative).

Fixing a claim turns the gain into a budget, set against the bits the claim needs. If a claim has probability $p$ with the full context and $q$ after a deletion, the deletion supplies $I_G\ge d_2(p\|q)$ bits, where $d_2$ is the binary divergence, and lifting the claim from $q$ to a target $t>q$ or higher needs at least $d_2(t\|q)$ bits (Corollary~\ref{cor:sharp} in Appendix~\ref{sec:budget}). The 300 feature points supply 17.66 bits against a requirement of 16.93, and every confident feature claim falls without them, while the 300 matched points supply 0.001 bits against 0.006. An observing view supplies 1.62 bits against 1.27 and accounts for 2.4 times as many lost claims as a non-observing view (Table~\ref{tab:budget} in Appendix~\ref{sec:results}). With a binary claim and an evidence passage as the deleted part, the budget is the Trace Budget adjudicator of \citet{chlon2026predictable}, and the law places it inside the decomposition: for one context, the bits the evidence supplies are the decoding loss of the memory rebuilt without it, measured against the full-context prediction. A language-model verifier scores the claim with its evidence present and removed, and the budget gap, the bits the claim requires minus the bits its evidence supplies, separates supporting from insufficient evidence with AUROC 0.91 for both Qwen2.5-7B and GPT-4.1-mini on VitaminC.

\section{Choosing between context and weights}\label{sec:choose}
The two channels write differently, a transformer's cache as a weight update whose step sizes are recomputed for every query and a fine-tune as one fixed update (Appendices~\ref{sec:tune}--\ref{sec:kriging}), and they differ in what they can store.

\begin{proposition}[Capacity]\label{prop:capacity}
By data processing, the storage loss is at least $[I(Q;D)-I(D;Z_c)]_+$. A context memory rebuilt from the part $D_w$ of the context inside its window, with $D=(D_w,D_{\mathrm{ev}})$, has $\stor_A\ge I(Q;D_{\mathrm{ev}}\mid D_w)$, the information evicted (zero when the context fits); equality holds when the memory keeps its window, $I(Q;Z_A)=I(Q;D_w)$, as a transformer's cache does, since its layer-1 keys and values are generically injective in the tokens. A fine-tune with $P_{\mathrm{train}}$ trainable parameters stored at $b$ bits each has $I(D;\theta_T)\le b\,P_{\mathrm{train}}$.
\end{proposition}

Fine-tuning opens no space that grows with the data. A fine-tune writes into $bP_{\mathrm{train}}$ bits fixed before the data arrive. A full fine-tune, whose trainable parameters are all of the model's, writes into the bits that already hold the pretrained model: with $D$ independent of the pretraining data $D_0$, $I(\theta_T;D)+I(\theta_T;D_0)\le I(\theta_T;D,D_0)\le H(\theta_T)\le bP_{\mathrm{train}}$, so the new data and the old share one budget. A context memory is built afresh from each context and leaves the weights untouched. Because both channels face the same $G^\star$, Equation~\eqref{eq:law2} compares them exactly:
\begin{equation}\label{eq:crossover}
\E[\Gamma_A]-\E[\Gamma_B]=(\stor_B-\stor_A)+(\dec_B-\dec_A).
\end{equation}

\subsection{Ceilings and certificates}\label{sec:certificates}
Let the context hold $N$ examples which, like the query, are drawn independently from one task. The context memory reads the $W$ of them inside its window. Write $\KL_0=\KL(P^*_Q\|p_0)$ for the prior repair the two channels share, and let $C_B$ be any bound on the bits the fine-tune stores, $I(D;Z_B)\le C_B$, such as $bP_{\mathrm{train}}$ or the noisy-step bound of Proposition~\ref{prop:noisy} in Appendix~\ref{sec:noisy}.

\begin{proposition}[Choosing the channel]\label{prop:choose}
Under Assumption~\ref{ass:clean}, with the context memory rebuilt from the window $D_w$:
\begin{enumerate}[label=(\roman*)]
\item $\E[\Gamma_A]\le I(Q;D_w)+\KL_0$ and $\E[\Gamma_B]\le\min\{I(Q;D),C_B\}+\KL_0$.
\item Learning in context wins whenever $\E[\Gamma_A]$ exceeds $\min\{I(Q;D),C_B\}+\KL_0$, and fine-tuning wins whenever $\E[\Gamma_B]$ exceeds $I(Q;D_w)+\KL_0$.
\item If the context fits the window and the context memory keeps it, as a transformer's cache does (Proposition~\ref{prop:capacity}), then $\E[\Gamma_B]-\E[\Gamma_A]\le\dec_A$.
\item For $N\ge W$, $\E[\Gamma_A]$ does not depend on $N$, so a fine-tune whose capacity bound satisfies $C_B+\KL_0<\E[\Gamma_A]$ at every $N$ loses at every $N\ge W$.
\end{enumerate}
\end{proposition}

Each certificate in (ii) needs one channel's measured gain and the other's ceiling, so when it fires it settles the choice without running the other channel. When the context fits and the memory keeps it, part (iii) caps what fine-tuning can add at $\dec_A$, the decoding loss in context. Turned around, any fine-tune's advantage over learning in context is a lower bound on $\dec_A$. That makes it a test of whether the model uses what its memory holds, and the bound holds context by context (Appendix~\ref{app:choose}). Part (iv) turns the capacity ceiling into a statement about data: a fine-tune whose capacity bound, plus the prior repair, falls short of the gain in context never catches up, however much data it sees.

Each tool covers its own regime. The fine-tune certificate needs a context that overflows the window: when the context fits, the in-context ceiling $I(Q;D_w)+\KL_0$ equals $G^\star$, which no channel can exceed, and part (iii) takes over. The certificate for learning in context can fire only when $C_B<I(Q;D_w)$, and it takes any valid storage bound: an adapter's $kb$ bits (Section~\ref{sec:decisiontest}), or, for a fine-tune with noisy steps, the bound of Proposition~\ref{prop:noisy}. The transfer bound of Proposition~\ref{prop:transfer} replaces the capacity ceiling altogether. The certificates need the information terms and the prior repair, which are exact for designed distributions of contexts (Section~\ref{sec:decisiontest}); part (iii), the learning curves, the storage budget and the anytime-valid test need neither, so Section~\ref{sec:domains} runs them on GPT-2, tables and airfoil pressure, where those terms are unknown.

\subsection{Choosing without the true distribution}\label{sec:curves}
Let the context consist of units $d_1,\dots,d_n$, such as examples, observations, records or question--answer pairs, write $D_m=(d_1,\dots,d_m)$, and let the query $Q$ be a further unit. The learning curve of channel $c$ is $L_c(m)=\E\,\ell_c(Q\mid D_m)$, and fine-tuning on $n$ units beats a context memory holding $N$ units exactly when $L_B(n)<L_A(N)$.

\begin{proposition}[Prequential learning curves]\label{prop:preq}
Let $d_1,\dots,d_n$ and $Q$ be exchangeable, and let each channel build its memory after $m$ units from those units and from randomness independent of the data: the context memory by reading them through its window, and the fine-tune by one online update per unit, $\theta_m=U(\theta_{m-1},d_m,\xi_m)$. Then for every $j\le n$,
\begin{equation}\label{eq:preq}
\E\,\ell_c(d_j\mid d_{<j})=L_c(j-1),\quad \E\sum_{j=1}^{n}\ell_c(d_j\mid d_{<j})=\sum_{m=0}^{n-1}L_c(m).
\end{equation}
\end{proposition}

In context, the $j$-th term rebuilds the memory from the units before unit $j$ and scores unit $j$; for a transformer whose window holds the context, one teacher-forced pass returns every term. For the online fine-tune, $\ell_B(d_j\mid d_{<j})=\ell_{\theta_{j-1}}(d_j)$ is the first-epoch loss on unit $j$. Each sum is a prequential code length of $D$ \citep{dawid1984prequential,rissanen1984universal,blier2018description}. A sum adds up the whole learning curve, and learning curves cross, so the winner at a given size comes from the terms at that size, averaged over orderings, or from held-out queries.

\begin{theorem}[Anytime-valid choice]\label{thm:eprocess}
Fix the task and the context $D$, build the context memory from $D$ and train any model on $D$, by fine-tuning or from scratch, and let $q_1,q_2,\dots$ be held-out queries drawn independently from the task. Mix the context memory's probability for each output symbol with $\varepsilon$ of the uniform distribution over the alphabet $V$, so that for queries of at most $L$ symbols $\Delta_j=\ell_B(q_j)-\ell_A(q_j\mid D)\ge-c$ with $c=L\log_2(|V|/\varepsilon)$. Under the hypothesis $H_0$ that the trained model is no worse than this context memory on the task, $\E[\Delta_j]\le0$, the process
\begin{equation}\label{eq:eprocess}
E_t=\prod_{j\le t}\Big(1+\lambda_j\frac{\Delta_j}{c}\Big),\quad \lambda_j\in[0,1)\ \text{chosen from }D\text{ and }q_{<j},
\end{equation}
is a nonnegative supermartingale with $E_0=1$, and $\Pr(\sup_tE_t\ge1/\alpha)\le\alpha$.
\end{theorem}

The process is a test by betting \citep{ramdas2023gametheoretic,waudbysmith2024betting}, and its guarantee is Ville's inequality \citep{ville1939}. Whenever the test stops, reaching $1/\alpha$ rejects $H_0$ at level $\alpha$ and calls learning in context the winner on this task with this context. Exchanging the channels gives the test in favour of the trained model, whether fine-tuned or trained from scratch. Continuous outputs are discretised first, so the test applies to any memory.

\paragraph{Storage budget.} Fine-tuning replaces the context by a fixed state. Let $S$ be any such state built from $D$, with randomness independent of the sampled in-context outputs: fine-tuned weights, an adapter, or a compiled or compressed replacement of the context memory. Let any decoder $p_S(\cdot\mid x)$ read the state, and let a workload of $K$ inputs $x_1,\dots,x_K$ have in-context outputs $\hat Y_j\sim p_A(\cdot\mid D,x_j)$, sampled independently given $D$.

\begin{theorem}[Storage budget]\label{thm:storage}
\begin{equation}\label{eq:storage}
\sum_{j=1}^{K}\E\,\KL\big(p_A(\cdot\mid D,x_j)\,\big\|\,p_S(\cdot\mid x_j)\big)\;\ge\;I(\hat Y_{1:K};D)-I(D;S),
\end{equation}
and $I(D;S)\le H(S)$ when $S$ takes finitely many values.
\end{theorem}

A state that reproduces the context memory on the workload to within $\varepsilon$ bits in total stores at least $I(\hat Y_{1:K};D)-\varepsilon$ bits about $D$, so no fine-tune whose capacity bound is below $I(\hat Y_{1:K};D)-\varepsilon$ can reproduce it that closely. The information can be estimated from the model alone, by resampling the content of $D$ and scoring the in-context outputs under every variant (Appendix~\ref{app:Ihat}).

\section{Testing the choice}\label{sec:evidence}
\subsection{Certificates with exact information terms}\label{sec:decisiontest}
\begin{figure}[t]
\centering
\includegraphics[width=\columnwidth]{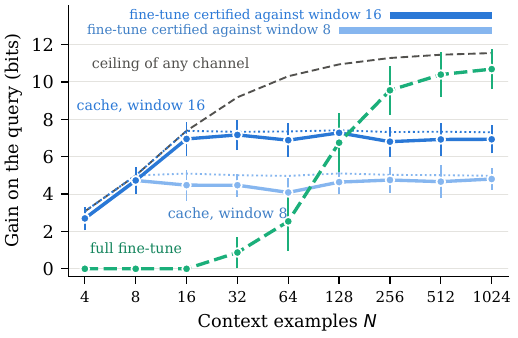}
\caption{Cache against fine-tuning on a trained transformer with exact information terms: gain on a 16-example query against the number of context examples. Dotted: ceilings $I(Q;D_w)+\KL_0$ of the caches with windows 8 and 16; dashed: ceiling $I(Q;D)+\KL_0$ of any channel; bars: sizes at which the fine-tune's gain passes the cache's ceiling, which certifies fine-tuning. 95\% intervals over 64 tasks.}
\label{fig:decision}
\end{figure}

Proposition~\ref{prop:choose} is tested where the information terms are exact. A three-layer transformer with 104,194 parameters is trained to learn linear-Gaussian regression tasks in context ($d=8$, signal-to-noise ratio $\rho=4$; Appendix~\ref{sec:closedforms}). A query is a block of 16 teacher-forced examples, and the context holds $N$ further examples of the task, $N=4$ to 1,024. The cache reads the last $W=8$ or $W=16$ examples; the full fine-tune trains all weights with a recipe tuned for each $N$ on separate tasks; and adapters train $k$ pseudo-labels quantised to $b$ bits, so they store at most $C_B=kb$ bits. A certificate is issued when its margin, the measured gain minus the other channel's ceiling, has a 95\% interval over 64 tasks above zero, and the measured winner is the channel whose paired gain difference has a 95\% interval excluding zero, otherwise a tie (Appendices~\ref{sec:decisionfull} and~\ref{app:decision}).

Of the 64 distinct comparisons between a cache and a fine-tune, 25 are certified, and every certified call agrees with the measured winner (Table~\ref{tab:decision} in Appendix~\ref{sec:decisionfull}). The cache's gain stops growing once the context passes the window and stays within 0.9 bits of its ceiling, while the fine-tune's rises from 0.9 bits at $N=32$ to 10.7 bits at $N=1{,}024$, against a ceiling of 11.5 bits for any channel (Figure~\ref{fig:decision}). The fine-tune beats the cache from $N=128$ with the window of 8 and from $N=256$ with the window of 16, and from there on its gain passes the cache's ceiling, which certifies fine-tuning against every cache with that window. When the context fits the window, every fine-tuning schedule lowered the gain on the tuning tasks, so the tuned recipe leaves the weights unchanged and the cache wins by 2.7 to 7.0 bits. There, Proposition~\ref{prop:choose}(iii) caps what any fine-tune could add at the cache's decoding loss, estimated at 0.27 to 0.43 bits. Even with the prior repair of 0.37 bits added, the capacity of the 2-bit and 4-bit adapters falls short of the gain of the cache with the window of 16, so by Proposition~\ref{prop:choose}(iv) they lose at every $N\ge16$. They lose at every size tested, and the certificate for the cache fires against both (Figure~\ref{fig:decisionfull}b in Appendix~\ref{sec:decisionfull}).

Across TabPFN, an occupancy network and a stereo network, with contexts drawn from designed distributions, 36 of 96 certificates are issued and all 86 checks against measured winners agree (Appendix~\ref{app:certnative}). TabPFN reading 1,024 rows is certified against every memory of at most 8 bits, and logistic regression trained on the same rows against every memory confined to the last 16.

\subsection{Choosing without the true distribution}\label{sec:domains}
\begin{figure*}[t]
\centering
\includegraphics[width=\textwidth]{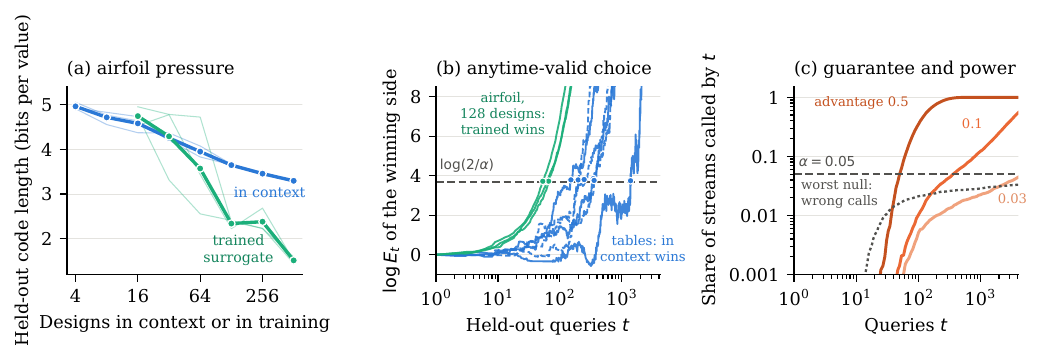}
\caption{The choice across domains. (a) Airfoil pressure: held-out code length per pressure value against the number of designs that the in-context learner reads or the surrogate trains on, mean over three seeds (thin lines: seeds); the in-context learner codes better with few designs and the surrogate from 128 on. (b) $\log E_t$ of the test that calls the winner, in each seed's query order, fixed in advance: TabPFN over XGBoost on phoneme (solid) and spambase (dashed), and the surrogate over the in-context learner on airfoil pressure with 128 designs; dots mark the calls. (c) Share of 10,000 simulated streams called by query $t$: correct calls for advantages of 0.5, 0.1 and 0.03 bits per query, and wrong calls under the worst of three nulls, against $\alpha=0.05$.}
\label{fig:domains}
\end{figure*}

We run Theorem~\ref{thm:eprocess} in both directions at level 0.025 each, so that a call errs with probability at most 0.05, on two choices outside language, each against a model trained from scratch (Appendices~\ref{app:domainsfull} and~\ref{app:domains}). On tables, TabPFN-v2 \citep{hollmann2025tabpfn} reads 128 labelled rows in its context, XGBoost \citep{chen2016xgboost} trains on the same rows, and the queries are 4,096 and 4,079 fresh rows of the OpenML phoneme and spambase datasets \citep{vanschoren2013openml}. On airfoil pressure, from 1,000 AirfRANS simulations \citep{bonnet2022airfrans}, an in-context learner predicts the 64 surface pressure values of a new design from the pressure fields of its support designs, weighted by their similarity to it, and a two-layer MLP surrogate trains on the same 128 designs. Both calibrate their predictive spread on 128 further designs, and the queries are 254 designs from unseen geometry groups.

The test picks TabPFN in all three seeds on both tables, after 151 to 1,390 queries, and the trained surrogate in all three seeds on airfoil pressure, after 53 to 66. It reaches the same call in each of 200 random reorderings of every seed's queries (Figure~\ref{fig:domains}b). The test compares code lengths, which reward calibrated probabilities. TabPFN codes a label in 0.61 bits against XGBoost's 0.81 on phoneme, where XGBoost's labels are slightly more accurate (Table~\ref{tab:domainruns}), and in 0.33 against 0.56 on spambase; the surrogate codes a pressure value in 2.34 bits against the in-context learner's 3.65. The learning curves show where the airfoil call comes from (Figure~\ref{fig:domains}a): the in-context learner improves at every size tested, from 4.96 bits per value with 4 designs to 3.30 with 512, but the surrogate overtakes it by 32 to 128 designs, and the test follows the crossing.

The guarantee holds where it can be checked. On 10,000 simulated streams of score differences with a known mean, the test calls a winner within 4,096 queries in 3.25\%, 2.87\% and 3.31\% of streams under three nulls (symmetric, skewed and of changing variance) and in none under an exact tie, while an advantage of 0.5 bits per query is called correctly in every stream, after a median of 127 queries (Figure~\ref{fig:domains}c; Table~\ref{tab:controls} in Appendix~\ref{app:domainsfull}).

On GPT-2 \citep{radford2019gpt2} the same tools apply unchanged\label{sec:gpt2} (Appendices~\ref{sec:gpt2full} and~\ref{app:gpt2}). Over 32 comparisons between its cache and an online fine-tune, on Shakespeare passages \citep{karpathy2015charrnn} and document recall, each run on 2,000 query streams, the test calls the better memory in 61,948 of the 64,000 streams and the worse one in 4, within its error limit. Four passages fit GPT-2's window and its cache keeps them, so by Proposition~\ref{prop:choose}(iii) the fine-tune's advantage on held-out passages, 0.35 to 0.55 bits per token, is a lower bound on GPT-2's decoding loss: its prediction falls at least that far short of the best prediction from its prompt.

\section{Related work}\label{sec:related}
Minimum description length identifies learning with compression \citep{rissanen1984universal,grunwald2007mdl}, and prequential codes score sequential forecasters \citep{dawid1984prequential}. Deep networks compress their training data well under prequential codes \citep{blier2018description}, the next-token loss that trains an in-context learner is a prequential code length \citep{elmoznino2025occam}, and language models are strong compressors \citep{deletang2024compression}. This account carries the view from the weights to the context of any model, and splits each code length saved into bits stored and bits used. The gain is pointwise usable information \citep{xu2020usable,ethayarajh2022vusable} with the family reduced to one frozen model and its memory, and Theorem~\ref{thm:law} splits it exactly into the Shannon information the memory stores, a prior repair and a decoding loss. Data valuation scores training examples by retraining or by influence approximations \citep{koh2017influence,ghorbani2019shapley,ilyas2022datamodels}, and context attribution scores parts of a prompt by ablation \citep{fernandes2021cxmi,cohenwang2024contextcite}. In context, deleting a part and rebuilding the memory gives its exact leave-one-out value at one rerun per part, and Corollary~\ref{cor:ablation} with a matched control splits that value into content and form.

In-context learning has been interpreted as implicit Bayesian inference \citep{xie2022explanation}, as a learned algorithm \citep{garg2022can}, as gradient descent or ridge regression run by attention \citep{vonoswald2023,akyurek2023learning,mahankali2024} and as implicit fine-tuning \citep{dai2023gpt}, and training, conversely, as attention over training patterns \citep{irie2022dual}. Information-theoretic analyses express its error through the information a context carries about a latent task \citep{jeon2024information}, and in-context learning now runs beyond language, as in prior-fitted networks for tables \citep{hollmann2025tabpfn}. Proposition~\ref{prop:cache} (Appendix~\ref{sec:cache}) writes softmax attention exactly as a weight write with query-dependent step sizes, using the context/prefix split of \citet{he2022unified} and the divided-difference identity of \citet{huang2026rope}, and the law prices that write and a fine-tune in the same bits.

Prompt-based fine-tuning often follows kernel dynamics \citep{jacot2018ntk,malladi2023kernel}, prefix tuning and low-rank adaptation write small states \citep{li2021prefix,hu2022lora}, and language models store a few bits of knowledge per parameter \citep{allenzhu2025capacity,morris2025memorize}; Proposition~\ref{prop:choose} turns any valid bound on what a channel stores into a certificate. Context distillation reproduces a context's predictions in the weights \citep{askell2021general,snell2022learning}, and Theorem~\ref{thm:storage} sets the floor of its divergence for a state that stores a given number of bits about the context. \citet{lampinen2025generalization} and \citet{mosbach2023fewshot} study in-context learning and fine-tuning side by side; Proposition~\ref{prop:choose} settles the choice from one measured gain when a certificate fires, and Theorem~\ref{thm:eprocess} settles it from held-out queries with anytime-valid error control, as a comparison of forecasters by betting \citep{henzi2022valid,choe2024comparing}. Appendix~\ref{app:related} discusses each line in full.

\section{Conclusion}
Measured in code length, what data give a pretrained model obeys one law in expectation, whether they sit in its context or in its weights: bits stored, plus prior repair, minus decoding loss. Its ceilings certified the choice between context and weights on a trained transformer, TabPFN, an occupancy network and a stereo network, and every certificate was correct; its anytime-valid test made the choice on tables, airfoil pressure and GPT-2 within its error limit. The same evaluator valued the parts of a context in occupancy, stereo and kriging models and recovered a published evidence adjudicator for two language-model verifiers. The law gives a compression account of learning from context.

\section*{Impact Statement}
This paper gives a measured basis for deciding whether new data should go into a model's context or into its weights, and a test of whether a prediction rests on the evidence supplied with it. The evidence adjudicator flags claims that their cited evidence does not support, which bears on the reliability of retrieval-augmented and fact-checking systems. The methods measure what a context supplies and add no new capability to the models they evaluate.

\bibliography{refs}
\bibliographystyle{plainnat}

\newpage
\appendix
\onecolumn
\section{Further results}\label{app:further}
This appendix gives the results that the main text uses or points to: the deleted-evidence lemma, the token-level form of the gain, the two writes, capacity under noisy steps, the lift budget, the estimator of the storage budget, the transfer integral, and a model in which every term is closed form. Proofs are in Appendix~\ref{app:proofs}.

\subsection{Deleted-evidence exclusion}\label{app:lemma}
\begin{lemma}[Deleted-evidence exclusion]\label{lem:exclusion}
Under Assumption~\ref{ass:clean}, two contexts that share the remaining part $R$ induce the same law for the rebuilt memory $Z_R$, whatever was deleted. In particular $G\perp\!\!\!\perp Z_R\mid R$, provided the exogenous randomness is conditionally independent of $G$ given $R$.
\end{lemma}

The rebuilt memory carries no trace of the deleted evidence, so a deletion is valued at what the evidence adds beyond the rest of the context, the query and the weights, just as held-out validation values a training example at what the rest of the training set does not already supply. The proof applies the fixed transition kernels step by step (Appendix~\ref{app:exclusion}).
\subsection{Where the gain comes from, token by token}\label{sec:token}
\begin{theorem}[Lift identity]\label{thm:token}
At a query position, let $p$ be the base next-token distribution and $p'=p\,e^{\Delta z}/\mathcal Z$ the channel's, where $\Delta z$ is the logit change the channel causes. The teacher-forced token $q$ gains, in nats,
\begin{equation}\label{eq:lift}
\log\frac{p'(q)}{p(q)}=\underbrace{\Delta z_q-\E_p[\Delta z]}_{\lambda\ \text{(centred lift)}}-\KL(p\,\|\,p').
\end{equation}
\end{theorem}

\begin{proof}
$\log\mathcal Z=\E_p\Delta z+\KL(p\|p')$.
\end{proof}

Hoeffding's lemma gives $0\le\KL(p\|p')\le\range(\Delta z)^2/8$, so the per-token gain $\Gamma_t$ obeys
\begin{equation}\label{eq:sandwich}
\lambda_t-\tfrac18\range(\Delta z_t)^2\;\le\;\Gamma_t\;\le\;\lambda_t.
\end{equation}
Over the query, $\sum_t\lambda_t>0$ is necessary for $Q$ to become more compressible, and $\sum_t\lambda_t>\sum_t\range(\Delta z_t)^2/8$ is sufficient. In expectation over the true next token,
\begin{equation}\label{eq:liftexp}
\E[\Gamma_t]=\langle P^*_t-p_t,\Delta z_t\rangle-\KL(p_t\|p'_t)=\KL(P^*_t\|p_t)-\KL(P^*_t\|p'_t).
\end{equation}
The cost $\KL(p\|p')$ grows with the size of the move, which is how a context can move predictions far more than a gradient step does and still gain little. To second order, with $h^*=\log(P^*/p)$ and the Fisher inner product $\langle a,b\rangle_p=\mathrm{Cov}_p(a,b)$,
\begin{equation}\label{eq:second}
\E[\Gamma_t]\approx\langle h^*,\Delta z\rangle_p-\tfrac12\|\Delta z\|_p^2 .
\end{equation}
To this order, the gain is positive when $\cos_p(h^*,\Delta z)>\|\Delta z\|_p/(2\|h^*\|_p)$, and a channel that moves logits along a direction $h$ captures at most $\cos^2_p(h,h^*)$ of the available $\KL(P^*\|p)$.

\subsection{The fine-tune}\label{sec:tune}
\begin{proposition}[Fine-tuning gain]\label{prop:tune}
Under gradient flow on $\ell(D)$ at rate $\eta$, in nats,
\begin{equation}\label{eq:kernel}
\begin{gathered}
\Gamma_B(Q)=\eta\int_0^T K_\nu(Q,D)\,d\nu,\\
K_\nu=\big\langle\nabla\log p_{\theta_\nu}(Q),\nabla\log p_{\theta_\nu}(D)\big\rangle=\sum_{t\in Q}\sum_{s\in D}r_{t}^{\top}J_{t}J_s^{\top}r_s ,
\end{gathered}
\end{equation}
where $\nu$ is training time, $t$ and $s$ index tokens of $Q$ and $D$, $r=e_{\text{token}}-p$ is the logit residual and $J=\partial z/\partial\theta$; for Adam, its preconditioner sits between the two gradients. For one discrete step $\theta_{k+1}=\theta_k+\eta g_D$, Taylor's theorem with integral remainder is exact:
\begin{equation}\label{eq:taylor}
\begin{aligned}
\Delta_k\log p(Q)={}&\eta\langle g_Q,g_D\rangle\\
&+\eta^2\!\int_0^1(1-a)\,g_D^{\top}\nabla^2\log p_{\theta_k+a\eta g_D}(Q)\,g_D\,da .
\end{aligned}
\end{equation}
\end{proposition}

The kernel $K_\nu$ is the empirical neural tangent kernel $J_tJ_s^\top$ \citep{jacot2018ntk}, contracted with the residuals of $Q$ and $D$. Let $\beta$ bound the absolute curvature of $\log p(Q)$ in the direction $g_D$ along the step. Then $Q$ becomes strictly more compressible if $\langle g_Q,g_D\rangle>\tfrac{\beta\eta}{2}\|g_D\|^2$, and it cannot improve if $\langle g_Q,g_D\rangle<-\tfrac{\beta\eta}{2}\|g_D\|^2$. Over $T$ steps, $\Gamma_B=\eta\sum_kK_k\pm\tfrac{\beta\eta^2}{2}\sum_k\|g_D(\theta_k)\|^2$. The first-order logit change is $\Delta z_t\approx\eta\sum_sJ_tJ_s^\top r_s$, and placing it in Equation~\eqref{eq:second} gives a one-step efficiency $\E[\Gamma_t]/\KL(P^*_t\|p_t)$ of at most $\cos^2_p(\Delta z_t,h^*)$, reached at the best step size.

\subsection{The cache}\label{sec:cache}
\begin{proposition}[The cache is a query-dependent weight write]\label{prop:cache}
Take any head with query vector $u$ and keys $k_i$, after rotary embedding \citep{su2024roformer}. Let $D$ be the context and $S$ the query's own prefix. Exactly,
\begin{equation}\label{eq:split}
\begin{gathered}
o(u)=\big(1-\alpha(u)\big)\,o_S(u)+\alpha(u)\,o_D(u),\\
\alpha(u)=\frac{\mathcal Z_D(u)}{\mathcal Z_D(u)+\mathcal Z_S(u)},\qquad \mathcal Z_X(u)=\sum_{i\in X}e^{\langle u,k_i\rangle}.
\end{gathered}
\end{equation}
With $\varphi(x)=(e^x-1)/x$ and an anchor query $u_0$, the context numerator is exactly
\begin{equation}\label{eq:divided}
\begin{aligned}
\sum_{i\in D}e^{\langle u,k_i\rangle}v_i={}&N_D(u_0)\\
&+\Big[\sum_{i\in D}e^{\langle u_0,k_i\rangle}\varphi\big(\langle u-u_0,k_i\rangle\big)\,v_ik_i^{\top}\Big](u-u_0),
\end{aligned}
\end{equation}
and the same holds for $\mathcal Z_D$ with $v_i$ replaced by $1$.
\end{proposition}

Equation~\eqref{eq:split} is the prefix decomposition of \citet{he2022unified}, with the context in the role of the prefix, and Equation~\eqref{eq:divided} is the divided-difference identity of \citet{huang2026rope}. The bracket is a weight matrix of rank at most $\min(|D|,d_h)$, and its per-example coefficients are recomputed for every query. Per head, the cache is a weight update whose step sizes depend on the query. A static weight edit, such as a fine-tune or a compiled replacement of the cache, writes the frozen-coefficient version, and the gap between the two is the transfer residual. Linear attention replaces the exponential by the identity, so its context numerator is exactly $\big(\sum_iv_ik_i^\top\big)u$: the write is $\sum_iv_ik_i^\top$, one unit step of gradient descent on $\tfrac12\sum_i\|v_i-Wk_i\|^2$ from $W=0$ \citep{vonoswald2023}. A logit perturbation $\delta$ changes every log-probability by at most $\range(\delta)$, so any static replacement of the cache costs at most $\range(\delta_t)$ nats per token, where $\delta_t$ is the logit discrepancy it induces.

\subsection{Attention without a cache}\label{sec:kriging}
A kriging learner predicts the value at a query site as a weighted sum of support values. The weights come from the agreement between the query and each support point, whitened by the agreement among the support points themselves through the Gram system, and constrained to sum to one. That is attention, with support descriptors as keys, support values as values and the Gram system in place of the softmax. A transformer stores its keys and values because its context is long and reused; a kriging learner recomputes its weights for each query because its support set is small. Deleting a support item and refitting is the same intervention as rebuilding a cache. The learner of Appendix~\ref{sec:native} also selects its own settings by leave-one-out prediction on its support set, so it runs held-out validation inside its own forward pass.

\subsection{Capacity under noisy steps}\label{sec:noisy}
\begin{proposition}[Noisy steps]\label{prop:noisy}
For noisy gradient steps with noise variance $\sigma_k^2$ and gradient norm at most $L$ \citep{pensia2018}, a fine-tune with $P_{\mathrm{train}}$ trainable parameters satisfies
\[
I(D;\theta_T)\le\sum_k\frac{P_{\mathrm{train}}}{2}\log\Big(1+\frac{\eta_k^2L^2}{P_{\mathrm{train}}\sigma_k^2}\Big).
\]
\end{proposition}

The evicted information is the quantity any compression of the context memory has to bound. For a context that fits the window of a memory that keeps it, as a transformer's cache does, deterministic, high-precision fine-tuning is generically injective in $D$ as well, so storage separates the two channels only for noisy, low-precision or low-dimensional updates; otherwise the difference between them is decoding loss. By Equation~\eqref{eq:crossover}, with $C_B$ the fine-tune's information capacity, learning in context wins whenever $[I(Q;D)-C_B]_+>\stor_A+\dec_A$, and, for a context memory that keeps its window, fine-tuning wins if and only if $I(Q;D_{\mathrm{ev}}\mid D_w)>\stor_B+\dec_B-\dec_A$.

\subsection{The lift budget}\label{sec:budget}
\begin{theorem}[Lift budget]\label{thm:budget}
For any event $E$, such as a correct answer, let $\hat Y_c\sim p_c(\cdot\mid Z_c)$ be the channel's output, $\pi_c(\cdot\mid D)=\E[p_c(\cdot\mid Z_c)\mid D]$ its law given $D$, and $\bar\pi_c=\E_D\,\pi_c(\cdot\mid D)$. Then
\begin{equation}\label{eq:liftbudget}
\begin{aligned}
\E_D\,d_2\big(\pi_c(E\mid D)\,\big\|\,p_0(E)\big)&\le I(\hat Y_c;D)+\KL(\bar\pi_c\,\|\,p_0)\\
&\le C_c+\KL(\bar\pi_c\,\|\,p_0),
\end{aligned}
\end{equation}
where $d_2(a\|b)=a\log_2\frac{a}{b}+(1-a)\log_2\frac{1-a}{1-b}$ and $C_c$ is the channel's capacity.
\end{theorem}

The probability lift a channel can buy is capped by the bits it writes, plus a drift term that does not depend on $D$. A memory computed deterministically from $D$, such as a feature grid or a key--value cache, has $\pi_c(\cdot\mid D)=p_c(\cdot\mid D)$. For a single context the budget is sharp. Write $b_t(q)=d_2(t\|q)$ for $q<t$ and $b_t(q)=0$ otherwise.

\begin{corollary}[Sharp lift budget]\label{cor:sharp}
Fix a claim $\cA$ with $p=p_c(\cA\mid D)$ and $q=p_c(\cA\mid R)$. Then $I_G\ge d_2(p\|q)$, with gap $p\,\KL\big(p_c(\cdot\mid D,\cA)\|p_c(\cdot\mid R,\cA)\big)+(1-p)\,\KL\big(p_c(\cdot\mid D,\cA^c)\|p_c(\cdot\mid R,\cA^c)\big)$. For $0<q<t<1$,
\begin{equation}\label{eq:sharp}
\min_{P:\,P(\cA)\ge t}\KL\big(P\,\|\,p_c(\cdot\mid R)\big)=b_t(q),
\end{equation}
attained by the law that keeps $p_c(\cdot\mid R)$ inside $\cA$ and $\cA^c$ and rescales their masses to $t$ and $1-t$. Consequently $p\ge t>q$ implies $I_G\ge b_t(q)$, and $I_G<b_t(q)$ implies $p<t$.
\end{corollary}

The minimiser is the information projection onto $\{P:P(\cA)\ge t\}$ \citep{csiszar1975idivergence}, and the budget is the Expectation-level Decompression Law of \citet{chlon2026predictable} applied to one context and its ablation.

\subsection{Estimating the storage budget}\label{app:Ihat}
The information on the right needs only the model. Resample the content of $D$ into $M$ variants $D^{(1)},\dots,D^{(M)}$, sample in-context outputs under each variant and score them under every variant; for $D$ drawn uniformly from the variants, and with $p_A(\hat Y_{1:K}\mid D)=\prod_jp_A(\hat Y_j\mid D,x_j)$,
\begin{equation}\label{eq:Ihat}
I(\hat Y_{1:K};D)=\frac1M\sum_{m=1}^{M}\E_{\hat Y\sim p_A(\cdot\mid D^{(m)})}\log_2\frac{p_A(\hat Y\mid D^{(m)})}{\frac1M\sum_{m'=1}^{M}p_A(\hat Y\mid D^{(m')})}.
\end{equation}
With the variants drawn from the data distribution, the expectation of the right side over the draws is a lower bound on $I(\hat Y_{1:K};D)$ that saturates at $\log_2M$ bits, so the estimate certifies the budget for states that store fewer bits than that. This is the information budget of \citet{chlon2026predictable}, computed over content-resampled datasets as that budget is computed over skeletons, with the bits the state stores about $D$ in the part of the bits-to-trust, and keeping $D$ in context in the part of abstention.

\subsection{The transfer integral}\label{app:transferint}
Write $\Gamma^{\mathrm{tr}}_B(t)=\ell_{\theta_0}(D)-\ell_{\theta_t}(D)$ for the code length that fine-tuning has removed from $D$ itself.

\begin{proposition}[Transfer integral]\label{prop:transfer}
Under gradient flow on $\ell(D)$ with any positive semidefinite preconditioner $P_t$, while $g_D^\top P_tg_D>0$,
\begin{equation}\label{eq:transfer}
\frac{d\Gamma_B(Q)}{d\Gamma^{\mathrm{tr}}_B}=r_t(Q)=\frac{g_Q^\top P_tg_D}{g_D^\top P_tg_D},\qquad\Gamma_B(Q)=\int_0^{\Gamma^{\mathrm{tr}}_B(T)}r\,d\Gamma^{\mathrm{tr}}_B,
\end{equation}
so that $|\Gamma_B(Q)|\le\sup_t|r_t(Q)|\,\Gamma^{\mathrm{tr}}_B(T)$, which for discrete $D$ is at most $\sup_t|r_t(Q)|\,\ell_{\theta_0}(D)$.
\end{proposition}

The ratio $r_t$ has no units, so the integral holds in bits. Under gradient flow, every bit removed from $D$'s code length transfers $r_t$ bits to $Q$: without compressing $D$, fine-tuning transfers nothing to $Q$, and along a run the integral splits the fine-tune's gain into the compression of $D$ and its transfer to the query. When the ratio stays within a band $|r_t(Q)|\le\rho$ along the run, the fine-tune gains at most $\rho\,\ell_{\theta_0}(D)$ bits on $Q$, so a fine-tune with $\rho\,\ell_{\theta_0}(D)<\E[\Gamma_A]$ cannot deliver what the context memory delivers, whatever its capacity, and a ratio that stays negative means interference. The ratio at initialisation, $r_0(Q)$, costs one backward pass per query and one for $D$ and fixes where the band starts; orthogonal gradients, as between a fact and its reversal, give $r_0=0$.

\subsection{A model where every term is exact}\label{sec:closedforms}\label{sec:closed}
Let $w\sim N(0,\tfrac{\bar\tau^2}{d}I)$, $x\sim N(0,I_d)$ and $y=w^\top x+\varepsilon$ with $\varepsilon\sim N(0,\sigma^2)$. A dataset holds $n$ examples; write $\gamma=d/n$ and $\rho=\bar\tau^2/\sigma^2$, and let the base predictive be the prior predictive of $y$ given $x$. A channel whose calibrated Gaussian predictive has excess risk $\mathcal E_c$ gains, in the proportional limit $n,d\to\infty$ with $\gamma$ fixed,
\begin{equation}\label{eq:gaussgain}
\E[\Gamma_c]=\tfrac12\log_2\frac{\sigma^2+\bar\tau^2}{\sigma^2+\mathcal E_c}.
\end{equation}

\begin{proposition}[Closed forms]\label{prop:closed}
(i) The ideal cache, the Bayes posterior predictive, gains $I(y_q;D\mid x_q)$, which in the proportional limit is Equation~\eqref{eq:gaussgain} with
\begin{equation}\label{eq:bayes}
\begin{gathered}
\mathcal E_A=\sigma^2\gamma\,s_\gamma(\kappa),\qquad\kappa=\gamma/\rho,\\
s_\gamma(\kappa)=\frac{-(\kappa+1-\gamma)+\sqrt{(\kappa+1-\gamma)^2+4\gamma\kappa}}{2\gamma\kappa}.
\end{gathered}
\end{equation}
(ii) One linear-attention layer, with its step size learned in pretraining, has, exactly at finite $n$ and $d$ with $\tau^2=\bar\tau^2/d$,
\begin{equation}\label{eq:la}
\mathcal E_{LA}=\tau^2d\,\frac{\tau^2(d+1)+\sigma^2}{\tau^2(n+d+1)+\sigma^2}.
\end{equation}
(iii) A fine-tune of $T$ gradient-descent steps of size $\eta$ from zero has, in the proportional limit,
\begin{equation}\label{eq:gd}
\begin{aligned}
\mathcal E_B(T,\eta)={}&\bar\tau^2\sum_{k=0}^{2T}\tbinom{2T}{k}(-\eta)^kM_k\\
&+\sigma^2\gamma\sum_{k,l=1}^{T}\tbinom Tk\tbinom Tl(-1)^{k+l}\eta^{k+l}M_{k+l-1},
\end{aligned}
\end{equation}
with $M_0=1$ and $M_k=\sum_{j=0}^{k-1}\frac{1}{j+1}\binom kj\binom{k-1}{j}\gamma^j$, the moments of the Marchenko--Pastur law \citep{marchenko1967}.
\end{proposition}

\begin{figure}[t]
\centering
\includegraphics[width=0.6\textwidth]{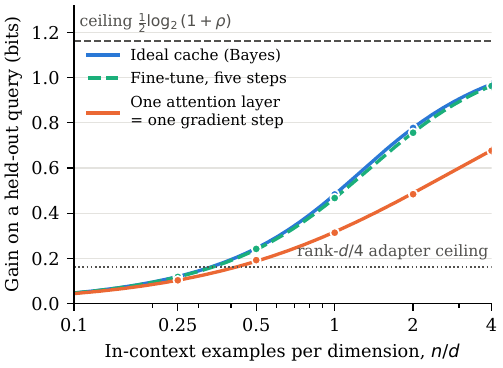}
\caption{Gain on a held-out query in the linear-Gaussian model at $\rho=4$. Lines are the closed forms of Proposition~\ref{prop:closed}; dots are simulations at $d=200$, 300 repeats each. One attention layer and one gradient step coincide, and five gradient steps nearly reach the ideal cache. The grey dashed line is the ceiling of every channel and the grey dotted line is the ceiling of a rank-$d/4$ adapter.}
\label{fig:closed}
\end{figure}

Equations~\eqref{eq:gaussgain}--\eqref{eq:gd} are the learning curves of Section~\ref{sec:curves} in closed form, and they give four results (Figure~\ref{fig:closed}).
\begin{itemize}
\item \textbf{One step equals one layer.} One gradient step at its best size has excess risk exactly $\mathcal E_{LA}$ at every $n$ and $d$, and in the proportional limit the best size is $\eta^*=\bar\tau^2/(\bar\tau^2(1+\gamma)+\sigma^2\gamma)$, so $\mathcal E_B(1,\eta^*)=\mathcal E_{LA}$. This turns the optimality of one gradient step for one layer of linear attention \citep{mahankali2024} into an identity of excess risks, and so of bits.
\item \textbf{Ordering.} $\mathcal E_A\le\mathcal E_B(T,\eta^*_T)\le\mathcal E_{LA}$, where $\eta^*_T$ is the best step; at $\rho=4$ this holds for every $T$ from 1 to 10 and every $n/d$ from 0.1 to 10.
\item \textbf{Adapter ceiling.} A $k$-dimensional adapter with a frozen random down-projection never exceeds $\tfrac12\log_2\frac{1+\rho}{1+(1-k/d)\rho}$. At $\rho=4$ and $k=d/4$, that is 14\% of the cache's ceiling $\tfrac12\log_2(1+\rho)$.
\item \textbf{Crossover.} At $\rho=4$, an ideal cache holding $N$ examples is matched by a fine-tune on $n^*$ examples, with $n^*/N=1.01$--$1.06$ for five or more steps and $n^*/N=1.2$--$3.8$ for one step, over $N/d$ from 1/4 to 4.
\end{itemize}
Simulation at $d=200$, with 300 repeats per point, reproduces the closed forms: over the points of Figure~\ref{fig:closed}, the largest gap between the two is 0.005 bits.

\section{Proofs}\label{app:proofs}

\subsection{Deleted-evidence exclusion}\label{app:exclusion}
Fix the remaining inputs $R=r$. Let $T_0$ denote the initialised state and write the successive transitions of the rebuilt execution as
\begin{equation}\label{eq:transitions}
T_0=\phi_0(r,U_0),\qquad T_{j+1}=\phi_{j+1}(r,T_0,\dots,T_j,U_{j+1}).
\end{equation}
The random sequence $U$ has a conditional law that depends on $r$ but not on the deleted content $G$. By induction, the conditional law of each transcript prefix $(T_0,\dots,T_j)$ depends only on $r$, and is therefore identical for every value of $G$ compatible with $r$. The same holds for a variable-length transcript when stopping is a measurable function of the available transcript and randomness. This proves Lemma~\ref{lem:exclusion}.

A state cached from the full context is admissible only if it is independent of the removed contents given the remaining inputs. In a transformer, the entries of the deleted span and every entry computed after it are recomputed. A cached geometry-only state qualifies when its arguments are unchanged; a chart or hyperparameter choice computed from deleted geometry does not. Predictions reconstructed from the model's parameters or the remaining context are always admissible. The conditional independence statement concerns the provenance of the rebuilt execution and makes no assumption that context examples are statistically independent.

\subsection{The law and its corollaries}\label{app:law}
\paragraph{Theorem~\ref{thm:law}.} The pointwise identity in the proof holds wherever the densities are positive, and its three terms have expectations $I(Q;Z_c)$, $\KL(P^*_Q\|p_0)$ and $\E_Z\KL(P^*_{Q\mid Z}\|p_c(\cdot\mid Z))$ under $P^*$, which gives Equation~\eqref{eq:law} whenever the three terms are finite.

\paragraph{Equation~\eqref{eq:law2}.} By the chain rule, $I(Q;D,Z_c)=I(Q;D)+I(Q;Z_c\mid D)=I(Q;D)$, because $Q\perp Z_c\mid D$, and also $I(Q;D,Z_c)=I(Q;Z_c)+I(Q;D\mid Z_c)$. Hence $I(Q;Z_c)=I(Q;D)-I(Q;D\mid Z_c)$. Moreover $\E_D\KL(P^*_{Q\mid D}\|p_0)=\E\log_2\frac{P^*(Q\mid D)}{P^*(Q)}+\E\log_2\frac{P^*(Q)}{p_0(Q)}=I(Q;D)+\KL(P^*_Q\|p_0)$.

\paragraph{Corollary~\ref{cor:law}.} For a predictor that sees $D$, $\E[\Gamma_c]=\E\log_2\frac{P^*(Q\mid D)}{p_0(Q)}-\E\log_2\frac{P^*(Q\mid D)}{p_c(Q\mid D)}=G^\star-\E\,\KL(P^*_{Q\mid D}\|p_c(\cdot\mid D))$, with the expectation over $D$ and the memory, which gives (i) and, by subtraction, (ii). For (iii), data processing along $Q-\tilde D-Z_c$ gives $I(Q;Z_c)\le I(Q;\tilde D)=0$, and the decoding loss is nonnegative.

\paragraph{Corollary~\ref{cor:ablation}.} By Equation~\eqref{eq:ablation}, $\E[\Delta_G]=\E[\Gamma_c(Q;D)]-\E[\Gamma_c(Q;R)]$. Equation~\eqref{eq:law2} applied to each context gives $\E[\Gamma_c(Q;X)]=I(Q;X)+\KL(P^*_Q\|p_0)-\stor_X-\dec_X$ for $X\in\{D,R\}$; the prior-repair term is common to both and cancels, and the chain rule gives $I(Q;D)-I(Q;R)=I(Q;G\mid R)$, which is Equation~\eqref{eq:ablationlaw}. When $I(Q;Z_X)=I(Q;X)$, $\stor_X=0$. For a matched deletion $\tilde G$ the content term $I(Q;\tilde G\mid D\setminus\tilde G)$ is zero, which leaves the storage and decoding change.

\paragraph{The model's own valuation.} With $Q\sim p_c(\cdot\mid D)$, $\E[\Delta_G]=\E\log_2\frac{p_c(Q\mid D)}{p_c(Q\mid R)}=\KL(p_c(\cdot\mid D)\|p_c(\cdot\mid R))$. For a spatial output, the pair $(U,Y_U)$ with $U\sim\mu$ under both conditions has divergence $\KL(\mu\|\mu)+\int\KL(p^u_D\|p^u_R)\,\mu(du)$ by the chain rule, and the first term is zero. The construction needs no law over the whole field.

\paragraph{Memory and forecast.} Let $F_c=p_c(\cdot\mid Z_c)$, a function of $Z_c$. The chain rule for the divergence gives $\dec_c=\E\,\KL(P^*_{Q\mid Z_c}\,\|\,F_c)=I(Q;Z_c\mid F_c)+\E\,\KL(P^*_{Q\mid F_c}\,\|\,F_c)$, and $I(Q;Z_c)=I(Q;F_c)+I(Q;Z_c\mid F_c)$. Substituting both into Theorem~\ref{thm:law} gives $\E[\Gamma_c]=I(Q;F_c)+\KL(P^*_Q\|p_0)-\E\,\KL(P^*_{Q\mid F_c}\,\|\,F_c)$, the resolution and reliability of the logarithmic score \citep{brocker2009reliability,weijs2010kullback} measured against $p_0$; the term $I(Q;Z_c\mid F_c)$ cancels from this form and stays inside the decoding loss of the law.

\subsection{The lift identity}\label{app:token}
With $\mathcal Z=\E_p e^{\Delta z}$, $\log\frac{p'(q)}{p(q)}=\Delta z_q-\log\mathcal Z$ and $\KL(p\|p')=\E_p[\log\mathcal Z-\Delta z]=\log\mathcal Z-\E_p\Delta z$, which gives Equation~\eqref{eq:lift}. Hoeffding's lemma, $\log\E_pe^{X}\le\E_pX+\range(X)^2/8$, bounds $\KL(p\|p')$. Under the true next token, $\E_{P^*}\log\frac{p'}{p}=\E_{P^*}\Delta z-\log\mathcal Z=\langle P^*-p,\Delta z\rangle-\KL(p\|p')$, and the same expectation equals $\KL(P^*\|p)-\KL(P^*\|p')$. For small moves, $\KL(p\|p')=\tfrac12\mathrm{Var}_p(\Delta z)+O(\|\Delta z\|^3)$ and $\langle P^*-p,\Delta z\rangle=\E_p[(P^*/p-1)\Delta z]=\mathrm{Cov}_p(h^*,\Delta z)+O(\|h^*\|^2\|\Delta z\|)$, which gives Equation~\eqref{eq:second}. Along $\Delta z=ah$ the quadratic is maximised at $a=\langle h^*,h\rangle_p/\|h\|^2_p$ with value $\tfrac12\|h^*\|^2_p\cos^2_p(h,h^*)$, while $\KL(P^*\|p)=\tfrac12\|h^*\|^2_p$ to the same order.

\subsection{The fine-tune}\label{app:tune}
Under gradient flow $\dot\theta_\nu=\eta\nabla\log p_{\theta_\nu}(D)$, $\frac{d}{d\nu}\log p_{\theta_\nu}(Q)=\eta\langle\nabla\log p_{\theta_\nu}(Q),\nabla\log p_{\theta_\nu}(D)\rangle$, and integrating gives Equation~\eqref{eq:kernel}. The token-level form follows from $\nabla_\theta\log p_\theta(y)=J^\top r$ for a softmax output. For one step, apply $f(1)=f(0)+f'(0)+\int_0^1(1-a)f''(a)\,da$ to $f(a)=\log p_{\theta_k+a\eta g_D}(Q)$. If $|g^\top\nabla^2\log p\,g|\le\beta\|g\|^2$ along the step, the remainder lies within $\pm\tfrac{\beta\eta^2}{2}\|g_D\|^2$, which gives both step conditions and, summed over steps, the $T$-step bound.

\subsection{The cache}\label{app:cache}
Softmax attention over the union of the context keys $D$ and the prefix keys $S$ gives $o(u)=\frac{N_S(u)+N_D(u)}{\mathcal Z_S(u)+\mathcal Z_D(u)}$ with $N_X(u)=\sum_{i\in X}e^{\langle u,k_i\rangle}v_i$, and dividing the two parts by their own normalisers gives Equation~\eqref{eq:split}. For each key, $e^{\langle u,k\rangle}=e^{\langle u_0,k\rangle}e^{x}$ with $x=\langle u-u_0,k\rangle$, and $e^x=1+\varphi(x)\,x$, so $e^{\langle u,k\rangle}v=e^{\langle u_0,k\rangle}v+e^{\langle u_0,k\rangle}\varphi(x)\,v\,k^\top(u-u_0)$; summing over the context gives Equation~\eqref{eq:divided}. In linear attention the gradient of $\tfrac12\sum_i\|v_i-Wk_i\|^2$ at $W=0$ is $-\sum_iv_ik_i^\top$, so one unit step writes $\sum_iv_ik_i^\top$. For logits $z$ and a perturbation $\delta$, $\log\mathrm{softmax}(z+\delta)_q-\log\mathrm{softmax}(z)_q=\delta_q-[\mathrm{lse}(z+\delta)-\mathrm{lse}(z)]$, and the bracket lies between $\min\delta$ and $\max\delta$, so the change is at most $\range(\delta)$ in absolute value.

\subsection{Capacity and crossover}\label{app:capacity}
Data processing along $Q-D-Z_c$ gives $I(Q;Z_c)\le I(D;Z_c)$, so $\stor_c=I(Q;D)-I(Q;Z_c)\ge I(Q;D)-I(D;Z_c)$. A context memory is rebuilt from the window with randomness independent of $(D,Q)$, so data processing along $Q-D_w-Z_A$ gives $I(Q;Z_A)\le I(Q;D_w)$, and with $D=(D_w,D_{\mathrm{ev}})$ the chain rule gives $\stor_A=I(Q;D)-I(Q;Z_A)\ge I(Q;D)-I(Q;D_w)=I(Q;D_{\mathrm{ev}}\mid D_w)$, with equality when $I(Q;Z_A)=I(Q;D_w)$. For a transformer, $Z_A=\KV(D_w)$; if the layer-1 entries determine the tokens in the window, $\KV(D_w)$ determines $D_w$, and equality holds. With $P_{\mathrm{train}}$ parameters stored at $b$ bits each, $\theta_T$ has entropy at most $bP_{\mathrm{train}}$, which bounds $I(D;\theta_T)$. Equation~\eqref{eq:crossover} is the difference of two instances of Equation~\eqref{eq:law2}, which share $G^\star$. Since $\stor_B\ge[I(Q;D)-C_B]_+$ and $\dec_B\ge0$, the condition $[I(Q;D)-C_B]_+>\stor_A+\dec_A$ makes the difference positive; with $\stor_A=I(Q;D_{\mathrm{ev}}\mid D_w)$, the difference is negative exactly when $I(Q;D_{\mathrm{ev}}\mid D_w)>\stor_B+\dec_B-\dec_A$.

\subsection{Ceilings and certificates}\label{app:choose}
(i) Theorem~\ref{thm:law} with $\dec_c\ge0$ gives $\E[\Gamma_c]\le I(Q;Z_c)+\KL_0$. The context memory is rebuilt from $D_w$ with randomness independent of $(D,Q)$, so $Q-D_w-Z_A$ is a Markov chain and $I(Q;Z_A)\le I(Q;D_w)$. For the fine-tune, $Q-D-Z_B$ gives $I(Q;Z_B)\le I(Q;D)$ and $I(Q;Z_B)\le I(D;Z_B)\le C_B$. (ii) follows from (i). (iii) When the context fits and the context memory keeps it, $I(Q;Z_A)=I(Q;D)$, so $\stor_A=0$, and Equation~\eqref{eq:crossover} gives $\E[\Gamma_B]-\E[\Gamma_A]=\dec_A-\dec_B-\stor_B\le\dec_A$. (iv) For $N\ge W$ the window holds $W$ examples drawn independently from the task whatever $N$ is, so the joint law of $(D_w,Q)$, and with it $\E[\Gamma_A]$, does not depend on $N$; since $\min\{I(Q;D),C_B\}\le C_B$, (ii) applies at every $N\ge W$. Statements (i)--(iii) also hold conditionally on inputs that both channels observe, and (iv) holds for gains averaged over the inputs; Section~\ref{sec:decisiontest} evaluates them this way. Both channels observe the context, so conditioning on $D=d$ gives $\E[\Gamma_B-\Gamma_A\mid D=d]=\E\big[\KL(P^*_{Q\mid d}\,\|\,p_A(\cdot\mid Z_A))\mid D=d\big]-\E\big[\KL(P^*_{Q\mid d}\,\|\,p_B)\mid D=d\big]\le\E\big[\KL(P^*_{Q\mid d}\,\|\,p_A(\cdot\mid Z_A))\mid D=d\big]$, which is the decoding loss in context on $d$ when the context memory keeps $d$.

\subsection{Learning curves, the anytime-valid test and the budgets}\label{app:measure}
\paragraph{Proposition~\ref{prop:preq}.} By exchangeability, $(d_{<j},d_j)$ has the law of $(D_{j-1},Q)$. In both cases the channel's memory after $j-1$ units is the same measurable function of those units and of randomness independent of the data, so $\ell_c(d_j\mid d_{<j})$ and $\ell_c(Q\mid D_{j-1})$ have the same law and the same expectation $L_c(j-1)$. Summing over $j$ gives the second identity.

\paragraph{Theorem~\ref{thm:eprocess}.} Let $\mathcal F_j$ hold $D$, the two memories and $q_{\le j}$. With the task and $D$ fixed, the memories are fixed and the queries are independent and identically distributed, so $\E[\Delta_t\mid\mathcal F_{t-1}]=\E[\Delta_t]\le0$ under $H_0$. The mixed context memory gives $\ell_A(q\mid D)\le L\log_2(|V|/\varepsilon)$ and $\ell_B(q)\ge0$, so $\Delta_j\ge-c$ and each factor satisfies $1+\lambda_j\Delta_j/c\ge1-\lambda_j>0$. Because $\lambda_t$ is fixed by $\mathcal F_{t-1}$, $\E[E_t\mid\mathcal F_{t-1}]=E_{t-1}\big(1+\lambda_t\E[\Delta_t\mid\mathcal F_{t-1}]/c\big)\le E_{t-1}$, and Ville's inequality for nonnegative supermartingales \citep{ville1939} gives $\Pr(\sup_tE_t\ge1/\alpha)\le\alpha E_0=\alpha$. On the prequential increments, with $\mathcal F_j$ holding the first $j$ units and the channels' randomness, the same computation holds under the conditional hypothesis $\E[\Delta_j\mid\mathcal F_{j-1}]\le0$ for every $j$.

\paragraph{Theorem~\ref{thm:storage}.} Write $P_j(\cdot\mid D)=p_A(\cdot\mid D,x_j)$ and $P_j(\cdot\mid S)=\E[P_j(\cdot\mid D)\mid S]$. Since $\hat Y_j$ is sampled given $D$ independently of $S$, $\hat Y_j\perp\!\!\!\perp S\mid D$, and for any decoder that reads $S$,
\[
\E\,\KL\big(P_j(\cdot\mid D)\,\|\,p_S(\cdot\mid x_j)\big)=\E\,\KL\big(P_j(\cdot\mid D)\,\|\,P_j(\cdot\mid S)\big)+\E\,\KL\big(P_j(\cdot\mid S)\,\|\,p_S(\cdot\mid x_j)\big)\ge I(\hat Y_j;D\mid S).
\]
Given $(D,S)$ the outputs are independent, so $H(\hat Y_j\mid D,S,\hat Y_{<j})=H(\hat Y_j\mid D,S)$ while $H(\hat Y_j\mid S,\hat Y_{<j})\le H(\hat Y_j\mid S)$; hence $I(\hat Y_j;D\mid S,\hat Y_{<j})\le I(\hat Y_j;D\mid S)$, and the chain rule gives $\sum_jI(\hat Y_j;D\mid S)\ge I(\hat Y_{1:K};D\mid S)$. Finally $I(\hat Y_{1:K};D\mid S)=I(\hat Y_{1:K};D)+I(\hat Y_{1:K};S\mid D)-I(\hat Y_{1:K};S)=I(\hat Y_{1:K};D)-I(\hat Y_{1:K};S)$, and $I(\hat Y_{1:K};S)\le I(D;S)$ by data processing along $\hat Y_{1:K}-D-S$; when $S$ takes finitely many values, $I(D;S)\le H(S)$. The first term of the decomposition is the storage term and the second the decoding term; if $S$ determines $D$, the storage term is zero.

\paragraph{Equation~\eqref{eq:Ihat}.} With $D$ uniform on the variants, the marginal law of $\hat Y$ is $\bar p=\frac1M\sum_{m'}p_A(\cdot\mid D^{(m')})$, and $I(\hat Y;D)=\frac1M\sum_m\KL\big(p_A(\cdot\mid D^{(m)})\,\|\,\bar p\big)$, which is the right side. If instead $V=(D^{(1)},\dots,D^{(M)})$ is drawn independently from the data distribution and $U$ is uniform on $\{1,\dots,M\}$, then $D=D^{(U)}$ has the data distribution, $\hat Y$ depends on $(U,V)$ only through $D$, and the right side equals $I(\hat Y;U\mid V=v)$. Its expectation over $V$ is $I(\hat Y;U\mid V)=I(\hat Y;D)-I(\hat Y;V)\le I(\hat Y;D)$, and it never exceeds $H(U)=\log_2M$.

\paragraph{Proposition~\ref{prop:transfer}.} Under $\dot\theta_t=\eta P_tg_D$ with $g=\nabla_\theta\log p_\theta$, $\frac{d}{dt}\Gamma_B(Q)=\eta\,g_Q^\top P_tg_D$ and $\frac{d}{dt}\Gamma^{\mathrm{tr}}_B=\eta\,g_D^\top P_tg_D\ge0$. Dividing gives $r_t$, and since $\Gamma^{\mathrm{tr}}_B$ is nondecreasing, the change of variables gives the integral. Then $|\Gamma_B(Q)|\le\int|r|\,d\Gamma^{\mathrm{tr}}_B\le\sup_t|r_t|\,\Gamma^{\mathrm{tr}}_B(T)$, and $\Gamma^{\mathrm{tr}}_B(T)\le\ell_{\theta_0}(D)$ because a discrete code length is nonnegative. The ratio is unchanged when both gains are converted from nats to bits.

\subsection{The lift budget}\label{app:budget}
Given $D$, the output $\hat Y_c$ has law $\pi_c(\cdot\mid D)$, so as for $G^\star$, $\E_D\KL(\pi_c(\cdot\mid D)\|p_0)=I(\hat Y_c;D)+\KL(\bar\pi_c\|p_0)$. Data processing through $\one_E$ gives $d_2(\pi_c(E\mid D)\|p_0(E))\le\KL(\pi_c(\cdot\mid D)\|p_0)$, and the Markov chain $D-Z_c-\hat Y_c$ gives $I(\hat Y_c;D)\le I(Z_c;D)\le C_c$. This proves Theorem~\ref{thm:budget}.

For Corollary~\ref{cor:sharp}, let $P=p_c(\cdot\mid D)$, $P'=p_c(\cdot\mid R)$, $p=P(\cA)$ and $q=P'(\cA)$ with $0<p,q<1$. On $\cA$, $dP/dP'=(p/q)\,(dP_\cA/dP'_\cA)$ for the conditional laws $P_\cA$ and $P'_\cA$, and on $\cA^c$ the analogous ratio uses $(1-p)/(1-q)$. Integrating over the two sets gives the exact coarse-graining identity
\begin{equation}\label{eq:coarse}
\KL(P\|P')=d_2(p\|q)+p\,\KL(P_\cA\|P'_\cA)+(1-p)\,\KL(P_{\cA^c}\|P'_{\cA^c}),
\end{equation}
so $I_G\ge d_2(p\|q)$ with the stated gap. The derivative $\frac{\partial}{\partial r}d_2(r\|q)=\log_2\frac{r(1-q)}{q(1-r)}$ is positive on $(q,1)$, so for $q<t$ every law $\tilde P$ with $\tilde P(\cA)\ge t$ has $\KL(\tilde P\|P')\ge d_2(\tilde P(\cA)\|q)\ge d_2(t\|q)$. The law with density $t/q$ on $\cA$ and $(1-t)/(1-q)$ on $\cA^c$ relative to $P'$ integrates to one, keeps both conditional laws, and attains that value by making both conditional terms in Equation~\eqref{eq:coarse} zero. For $q<t$ and $p\ge q$, strict monotonicity gives $d_2(p\|q)\ge d_2(t\|q)$ if and only if $p\ge t$, which yields both implications.

\subsection{Closed forms}\label{app:closed}
\paragraph{Gain.} Given $x$, the prior predictive of $y$ is $N(0,\sigma^2+\bar\tau^2\|x\|^2/d)$, and $\|x\|^2/d\to1$. A calibrated Gaussian predictive has variance equal to its expected squared error, which for an error $w-\hat w$ independent of the query input is $\sigma^2+x^\top\E[(w-\hat w)(w-\hat w)^\top]\,x$, and this concentrates at $\sigma^2+\mathcal E_c$. The expected log-density of a Gaussian predictive whose variance $v$ equals its expected squared error is $-\tfrac12\log_2(2\pi e v)$, so the difference of the two tends to $\tfrac12\log_2\frac{\sigma^2+\bar\tau^2}{\sigma^2+\mathcal E_c}$, which is Equation~\eqref{eq:gaussgain}.

\paragraph{Ideal cache.} The posterior mean is $\hat w=(X^\top X+\lambda I)^{-1}X^\top y$ with $\lambda=\sigma^2/\tau^2$, and its error is $\lambda(X^\top X+\lambda I)^{-1}w-(X^\top X+\lambda I)^{-1}X^\top\varepsilon$. Its expected squared norm is $\sigma^2\,\E\,\mathrm{tr}(X^\top X+\lambda I)^{-1}=\sigma^2\gamma\,m(-\kappa)$, where $m$ is the Stieltjes transform of the empirical spectrum of $X^\top X/n$ and $\lambda/n=\gamma/\rho=\kappa$. In the proportional limit, $m(-\kappa)$ is the positive root of $\gamma\kappa m^2+(\kappa+1-\gamma)m-1=0$, which is $s_\gamma(\kappa)$. The base predictive is the prior predictive, so the prior repair is zero, and the Bayes predictive has zero decoding loss; by Theorem~\ref{thm:law} its gain is $I(y_q;D\mid x_q)$.

\paragraph{One layer, one step.} One step of size $\eta$ from zero gives $\hat w=\eta X^\top y/n$, with excess risk $\bar\tau^2-2\eta\bar\tau^2+\eta^2\,\E\|X^\top y/n\|^2$. Since $\E\,\mathrm{tr}((X^\top X)^2)=nd(n+d+1)$, $\E\|X^\top y/n\|^2=[\tau^2d(n+d+1)+\sigma^2d]/n$. Minimising over $\eta$ gives Equation~\eqref{eq:la} exactly at finite $n$ and $d$, the risk of one linear-attention layer with its best scalar step; in the proportional limit the minimiser is $\eta^*$ and the risk is $\bar\tau^2\gamma(\bar\tau^2+\sigma^2)/(\bar\tau^2(1+\gamma)+\sigma^2\gamma)$.

\paragraph{$T$ steps.} With $S=X^\top X/n$, $T$ steps from zero give $w-w_T=(I-\eta S)^Tw-[I-(I-\eta S)^T]S^{-1}X^\top\varepsilon/n$, where $[I-(I-\eta S)^T]S^{-1}$ denotes the polynomial $\sum_{k=1}^{T}\binom Tk(-1)^{k+1}\eta^kS^{k-1}$, defined for every $S$ including $d>n$. Its expected squared norm is $\tfrac{\bar\tau^2}{d}\E\,\mathrm{tr}(I-\eta S)^{2T}+\tfrac{\sigma^2}{n}\E\,\mathrm{tr}\big([I-(I-\eta S)^T]^2S^{-1}\big)$. Expanding both powers binomially and replacing $\tfrac1d\mathrm{tr}\,S^k$ by its limit $M_k$ gives Equation~\eqref{eq:gd}.

\paragraph{Adapter.} An adapter that moves the predictor only within a $k$-dimensional random subspace $U$ independent of $w$ leaves at least $\E\|P_{U^\perp}w\|^2=(1-k/d)\bar\tau^2$ of excess risk, and Equation~\eqref{eq:gaussgain} turns that floor into the ceiling.

\paragraph{Numbers.} The ordering, the crossover ratios and Figure~\ref{fig:closed} are computed from Equations~\eqref{eq:bayes}--\eqref{eq:gd} with the best step found by one-dimensional minimisation, and the simulations draw $w$, $X$ and $\varepsilon$ afresh in every repeat.

\section{Experiments in full}\label{app:experiments}
This appendix gives the full results summarised in Sections~\ref{sec:valuing} and~\ref{sec:evidence}, in the same order; the protocols are in Appendix~\ref{app:protocols}.

\subsection{One evaluator for any memory}\label{sec:design}
\begin{algorithm}[t]
\caption{Context ablation as held-out validation}\label{alg:eval}
\small
\begin{algorithmic}[1]
\Require Predictor $f_\theta$, query $x$, kept context $C_0$, context $C$, deletions $\cG$, observation rule $O$, target $t$
\State $P\gets\Call{PredictFresh}{f_\theta,x,C_0,C}$ \Comment{memory from $C$}
\State Fix claims and observation indices for every branch
\For{$G\in\cG$}
  \State $P_G\gets\Call{PredictFresh}{f_\theta,x,C_0,C\setminus G}$ \Comment{rebuilt}
  \State Check identical coordinates, readout and inference policy
  \State $I\gets\KL(PO\,\|\,P_GO)$ \Comment{Equation~\eqref{eq:info}}
  \For{each fixed claim $\cA$ on the observed output}
    \State $p\gets PO(\cA)$; $q\gets P_GO(\cA)$
    \State $J\gets d_2(p\|q)$; $b\gets b_t(q)$
    \State Record $(I,J,b,p,q)$ and whether $p\ge t>q$
  \EndFor
\EndFor
\end{algorithmic}
\end{algorithm}

\paragraph{One scorer, task-specific adapters.}
Algorithm~\ref{alg:eval} is the whole evaluator. For a spatial output with sites $u$ drawn from a fixed measure $\mu$, the bits per observation are $I^\mu_G=\int\KL(p^u_D\|p^u_R)\,\mu(du)$. An adapter supplies a prediction law or a claim probability, an admissible deletion and a common output observation; the scorer itself contains nothing architecture-specific. Every deletion rebuilds the memory from what remains (Assumption~\ref{ass:clean}): the occupancy network re-encodes its grid, the stereo network rebuilds its cost volume, the kriging learner refits on the remaining cars, and the language verifier recomputes its cache. The gain $I_G$ is computed whenever the law is available, and a claim-only adapter supplies $J$. The language experiments run the published Trace Budget adjudicator, the binary case of the same test (Appendix~\ref{sec:nlp}).

Claims are fixed before any deletion, either from geometry before any prediction (occupancy features) or from the full-context prediction (depth intervals and pressure tolerances), and no claim is re-centred afterwards. Deleting nothing reproduces the full prediction, and the depth and field adapters reproduce their native implementations exactly, including preprocessing and context-dependent state (Appendices~\ref{app:mvs} and~\ref{app:fields}). For a fixed query and $k$ single-item deletions the protocol runs $k+1$ predictions; group deletions add predictions, not scoring rules.

\paragraph{Matched controls and statistical units.}
Every experiment pairs the deletion of relevant evidence with a matched control of the same form (Table~\ref{tab:contract}): an equal-size deletion from unchanged geometry, a camera that does not observe the queried point, or donor evidence whose agreement with the claim is known from independent geometry. By Corollary~\ref{cor:ablation}, the control's gain is what a deletion of that form costs by itself, and the content of the relevant evidence accounts for everything beyond it. Mesh and depth references define whether evidence supports a claim, so relevance never rests on the source of the evidence, and reference pressures test the predictor's accuracy independently of its gains.

Spatial means are computed within each query population. Geometry results weight objects or constructed pairs equally; multi-view results average five reference views per scene and then the scenes; field results weight sites by area within a car and weight cars equally. Crossing counts are pooled and reported with their denominators. Scene-level intervals resample scenes, so pixel counts never act as sample sizes.

\subsection{Results across four memories}\label{sec:results}
\begin{figure*}[t]
\centering
\begin{minipage}[t]{0.455\textwidth}\centering
\includegraphics[width=\linewidth]{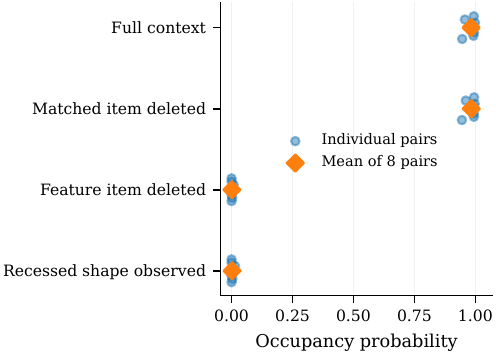}\\[1pt]
{\small (a) The feature grid of an occupancy network}
\end{minipage}\hfill
\begin{minipage}[t]{0.455\textwidth}\centering
\includegraphics[width=\linewidth]{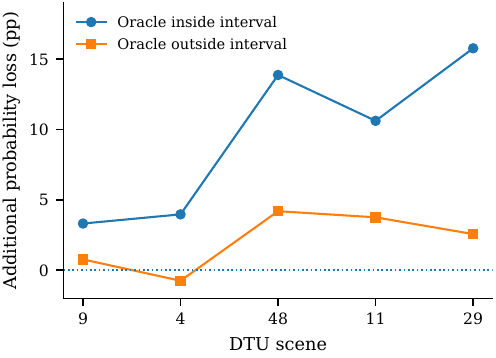}\\[1pt]
{\small (b) The cost volume of a stereo network}
\end{minipage}
\caption{Relevant evidence against a matched control of the same form. (a) Occupancy probability at the feature sites of eight constructed chair pairs, with the full context, after deleting 300 matched points from unchanged geometry, and after deleting the 300 points that observe the feature; the last row observes the recessed shape. Points are per-pair means over 1,024 sites and diamonds are means over pairs. (b) Extra loss of claim probability, in percentage points, from deleting a source view that observes the queried surface point rather than one that does not, by DTU scene; each point averages five reference views. Scene 9 was used to develop the protocol, and scenes 4, 48, 11 and 29 were drawn by seed before any of their predictions were seen. Claims whose interval misses the reference (oracle) depth are shown for comparison.}
\label{fig:evidence}
\end{figure*}

\begin{table*}[t]
\caption{One intervention, four memories. Counts are examples, objects, scenes or cars, which are the units of replication.}
\label{tab:contract}
\small
\begin{tabularx}{\textwidth}{@{}>{\raggedright\arraybackslash}p{0.105\textwidth}>{\raggedright\arraybackslash}p{0.14\textwidth}>{\raggedright\arraybackslash}p{0.135\textwidth}>{\raggedright\arraybackslash}p{0.12\textwidth}>{\raggedright\arraybackslash}p{0.15\textwidth}>{\raggedright\arraybackslash}X@{}}
\toprule
Task & Model and data & Memory built from the context & Deleted context & Output law and claim & Scale\\
\midrule
Controlled occupancy & Pretrained ConvONet; constructed chairs & Feature grid from the observed points & 300 observed surface points & Bernoulli occupancy; the site is occupied & 8 shape pairs; 1,024 feature sites per pair; relevant and matched deletions of equal size\\
Donor replacement & Same ConvONet; ShapeNet objects & Feature grid & Optional point-cloud observations & Bernoulli occupancy at fixed world coordinates & 16 objects; 8,492 fixed claims; independent donor-mesh labels\\
Multi-view depth & Pretrained MVSNet; DTU & Cost volume over the source views & Source image and camera & Categorical depth law; depth in a fixed interval & 5 scenes, 25 reference views; paired observing and non-observing deletions\\
Pressure fields & Kriging learner; DrivAerNet & Kriging weights, recomputed for each site & Support car (geometry and pressure) & Gaussian pressure law; pressure within a fixed tolerance & 8 query cars, 6 support cars each; 512 area-weighted sites\\
Language verification & Trace Budget adjudicator; VitaminC & Key--value cache & Evidence passage & Probability of the fixed claim & 800 examples; Qwen2.5-7B-Instruct and GPT-4.1-mini\\
\bottomrule
\end{tabularx}
\end{table*}

\subsubsection{A feature grid: 17.7 bits per query from a local shape feature}\label{sec:feature}
ConvONet encodes the observed points into a feature grid and decodes Bernoulli occupancy at any spatial query by reading the grid there \citep{peng2020convonet}. The grid is its memory of the context. We construct eight chair pairs that share their surrounding geometry and differ by a local raised feature versus a recess. Each condition keeps 300 common anchor points and nine optional items of 300 points: item 0 holds the observations that distinguish the pair, and item 1 holds a matched set from unchanged geometry. Query sites and claims (``this site is occupied'') are fixed from the constructed geometry before any prediction (Appendix~\ref{app:geometry}).

Deleting the distinguishing item and rebuilding the grid drops mean occupancy probability from 0.9827 to 0.0014 and costs 17.6642 bits per query on the full-context continuation (Table~\ref{tab:feature} and Figure~\ref{fig:evidence}a). Without the item the rebuilt grid predicts free space at the feature sites, as for the recess, and the 17.66 bits are what the item's observations save by overturning that confident prediction, priced site by site by Equation~\eqref{eq:info}. Deleting the same number of points from unchanged geometry leaves the probability at 0.9834 and costs 0.00111 bits. All 7,473 confident feature claims fall below 0.95 under the relevant deletion, against 70 under the matched control, and the full present-versus-absent contrast is positive in all eight pairs (mean 0.981).

Once the distinguishing item is deleted, the two shapes in a pair present identical inputs and receive identical predictions, which verifies the intervention end to end. Single-item gains also locate the relevant evidence: deleting the two items with the largest single-item gains (600 points) costs 17.58 bits, against 4.99 bits for a random pair of items.

\begin{table}[t]
\caption{Feature grid: eight pairs, 8,192 feature-site queries, $p=0.9827$ with the full feature-present context. Both deletions remove 300 points. The gain is averaged site by site. $X$ counts confident claims (7,473 with the full context) that fall below 0.95.}
\label{tab:feature}
\centering\footnotesize
\begin{tabular}{@{}lrrr@{}}
\toprule
Condition & Mean $q$ & Gain $I_G$ (bits) & $X$\\
\midrule
Feature item deleted & 0.001410 & 17.6642 & 7,473\\
Matched item deleted & 0.983387 & 0.00111 & 70\\
Recessed shape observed & 0.002191 & 17.4116 & ---\\
\bottomrule
\end{tabular}
\end{table}

\subsubsection{Donor evidence counts where its geometry agrees}
\begin{figure}[t]
\centering
\includegraphics[width=0.6\textwidth]{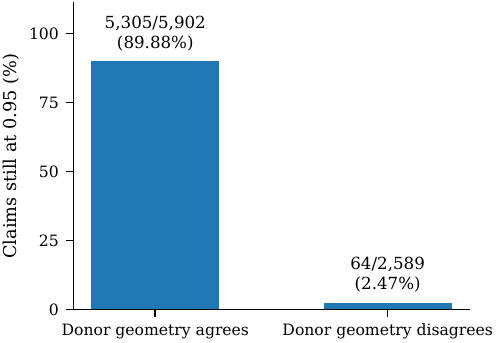}
\caption{Donor replacement on 8,492 fixed claims. Bars give the share of claims still at 0.95 when a same-category donor's observations replace the target's optional observations, split by whether the donor mesh agrees with the claim; rates are pooled over locations.}
\label{fig:donor}
\end{figure}

The natural-object experiments use 16 ShapeNet shapes, two in each of eight categories \citep{chang2015shapenet}. Donor replacement keeps 300 anchor points of the target and substitutes the optional observations of another object in the same category. We follow the 8,492 claims that the target's own optional observations lift above 0.95; with donor observations in their place, 5,369 of them stay confident. Independent donor meshes explain which ones:
\begin{center}\small
\begin{tabular}{@{}lrr@{}}
 & still at 0.95 / claims & rate\\
Donor agrees & 5,305/5,902 & 89.88\%\\
Donor disagrees & 64/2,589 & 2.47\%
\end{tabular}
\end{center}
In total, 98.81\% of the surviving confidence sits where the donor's geometry supports the claim (Figure~\ref{fig:donor}); equal-object rates are 78.34\% and 2.30\%, and one boundary-ambiguous claim is reported separately (Table~\ref{tab:donor}).

A coherent-object experiment replaces the whole observation set, anchors included. At interior coordinates occupied by the target but empty in the donor, mean occupancy probability is 0.9587 with the target's observations and 0.0351 with the donor's, averaging the eligible donors within each target and then the targets, and the contrast is positive for all 16 targets. Confidence tracks whether the evidence supports the claim at that coordinate, not which object the evidence came from.

\subsubsection{A cost volume: cameras that see the surface carry three times the bits}\label{sec:mvs}
MVSNet builds a cost volume from the reference and source views and reads a categorical distribution over depth hypotheses from it \citep{yao2018mvsnet}. We keep the reference image and camera and delete one of four source image--camera pairs, holding the depth hypotheses and each claim's interval fixed under native preprocessing. The experiment covers five DTU scenes \citep{jensen2014dtu} with five reference views each. Independent source depths classify each source view as observing the queried surface point, occluded, outside or behind the camera, or unknown. At each eligible query we delete one observing and one non-observing source, chosen by slot order rather than by model output, and views of unknown visibility are never used as non-observing. Each claim states that depth lies in a fixed interval around the full-context expected depth, and claims are grouped by whether the reference depth lies inside the interval, outside it, or near its boundary (Appendix~\ref{app:mvs}).

\begin{table}[t]
\caption{Cost volume: depth claims whose reference depth lies inside the fixed interval, 11,808 paired locations, of which 8,251 reach the target with the full context. Means weight scenes equally; crossing counts are pooled. Each deletion removes one source view.}
\label{tab:mvs}
\centering\footnotesize
\begin{tabular}{@{}lrr@{}}
\toprule
Deleted source view & Observing & Non-observing\\
\midrule
Full probability $p$ & 0.8952 & 0.8952\\
Ablated probability $q$ & 0.7552 & 0.8504\\
Gain $I_G$ (bits) & 1.6204 & 0.5452\\
Claim-level bits $J$ & 0.8391 & 0.1872\\
Target crossings & 1,840 & 760\\
Reverse crossings & 153 & 455\\
\bottomrule
\end{tabular}
\end{table}

Deleting an observing view and rebuilding the cost volume costs 1.6204 bits per query; deleting a non-observing view costs 0.5452 (Table~\ref{tab:mvs}). Through the depth claim the values are 0.8391 and 0.1872 bits, each below its gain as Corollary~\ref{cor:sharp} requires. Deleting the observing view also costs 9.52 more percentage points of claim probability, with the larger loss in all five scenes (Figure~\ref{fig:evidence}b; 95\% scene-bootstrap interval [5.04, 13.99]). On the four scenes added beyond the development scene, the effect is 11.06 points (interval [6.45, 14.83]). The effect is 4.5 times larger for correct depth intervals than for incorrect ones, whose extra loss is 2.11 points.

\subsubsection{Required and supplied bits}\label{sec:budgetresults}
Corollary~\ref{cor:sharp} turns every gain into a decision. For each deletion, Table~\ref{tab:budget} gives the bits a claim needs to return to 0.95 from its ablated probability, the bits the deleted context supplies, and how many claims actually cross the target. The supplied column is code length: when outcomes follow the full-context prediction, coding them with the rebuilt memory costs that many extra bits per query.

\begin{table}[t]
\caption{Bits required and supplied per query. Required: mean site-wise budget $b_{0.95}(q)$ after the deletion (zero when $q\ge 0.95$). Supplied: mean gain $I_G$. Claim: mean $J$. Crossings: confident claims that fall below 0.95. Means of site-wise values with the weighting of Tables~\ref{tab:feature} and~\ref{tab:mvs}; pressure uses native inference with claims at the native answer, weighting sites by area and cars equally.}
\label{tab:budget}
\centering\footnotesize
\begin{tabular}{@{}lrrrr@{}}
\toprule
Deletion & Required & Supplied & Claim & Crossings\\
\midrule
Feature item & 16.93 & 17.66 & 17.66 & 7,473/7,473\\
Matched item & 0.006 & 0.001 & 0.001 & 70/7,473\\
Observing view & 1.27 & 1.62 & 0.84 & 1,840/8,251\\
Non-observing view & 0.53 & 0.55 & 0.19 & 760/8,251\\
One support car & 0.74 & 0.23 & 0.04 & 4.3\% of area\\
\bottomrule
\end{tabular}
\end{table}

The feature item supplies 17.66 bits against a 16.93-bit requirement, and every confident feature claim falls without it; the matched control supplies 0.001 bits against 0.006. An observing view supplies 1.62 bits against a 1.27-bit requirement and accounts for 2.4 times as many lost claims as a non-observing view, which supplies 0.55 bits against 0.53. A single support car in the kriging learner supplies 0.23 bits against 0.74, and deleting it changes the decision on 4.3\% of the surface. In both tasks with a matched control, the relevant deletion supplies more bits and moves more claims across the target.

\subsubsection{Attention without a cache: native inference is part of the intervention}\label{sec:native}
The field predictor is a context-conditioned kriging learner for surface pressure on eight DrivAerNet query cars \citep{elrefaie2024drivaernet}, each with six support cars that pair geometry with CFD surface pressure. It attends to its support set without caching it (Appendix~\ref{sec:kriging}). The learner has no offline training: it selects its chart and regression settings from the support set by leave-one-car-out prediction, then transports a mean and an error-based spread to the query, giving a Gaussian law at every surface site (Appendix~\ref{app:fields}). Every deletion refits the learner on the remaining cars.

\begin{figure}[t]
\centering
\includegraphics[width=0.6\textwidth]{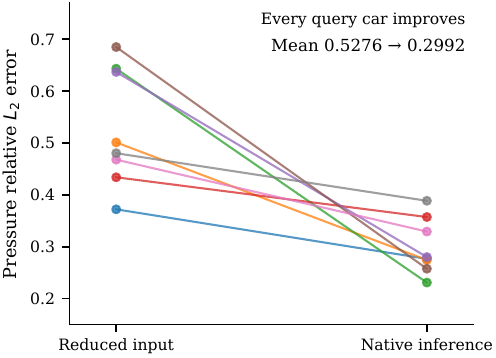}
\caption{Pressure error with the reduced input and with native inference. Lines connect the same query car at the same evaluation sites; the learner and the scorer are identical in both conditions.}
\label{fig:pressure}
\end{figure}

The learner's descriptors and bandwidth depend on the query geometry it receives. Passing only the 512 evaluation sites into inference therefore defines a different predictor from passing the full 32,000-point query and reading out the same 512 sites afterwards (Table~\ref{tab:fields}). Native inference cuts mean relative $L_2$ pressure error from 0.5276 to 0.2992, a 43.3\% reduction in the mean, with every one of the eight cars improving (Figure~\ref{fig:pressure}); the full query geometry alone gives 39.1\%. It also changes the measured gain: the mean single-car value falls from 0.79 to 0.23 bits, so the more accurate predictor relies less on any one support car. All 48 native single-car deletions carry positive gain, and the evaluated learner reproduces the original implementation's means, variances and selected configurations exactly in 168 comparisons.

\begin{table}[t]
\caption{Pressure prediction on eight fixed query cars and evaluation sites. The single-car gain averages the six deletions and eight cars. Reference pressure enters only the error.}
\label{tab:fields}
\centering\footnotesize
\begin{tabular}{@{}lrr@{}}
\toprule
 & Reduced input & Native inference\\
\midrule
Geometry points & 512 & 32,000\\
Chart anchors & 128 & 1,024\\
Leave-one-out points & 512 & 4,000\\
Mean relative $L_2$ error & 0.5276 & 0.2992\\
Mean single-car gain $I_G$ (bits) & 0.7939 & 0.2296\\
\bottomrule
\end{tabular}
\end{table}

Claims fix the native full-context answer at each site: pressure lies within 0.05 of it. Deleting a single car moves 4.3\% of the surface area across the 0.95 target, and a second claim bank fixed at the reduced-input answer gives 1.3\% (Table~\ref{tab:fieldbanks}).

\subsubsection{A key--value cache: the binary case is evidence adjudication}\label{sec:nlp}
With a binary claim and an evidence passage as the deleted span, the lift budget of Corollary~\ref{cor:sharp} is the Trace Budget adjudicator of \citet{chlon2026predictable}. A verifier scores the claim with the evidence present and with the cited evidence scrubbed. The adjudicator passes the claim when the evidence lifts it across the target: the full-context probability reaches 0.95 and the scrubbed probability does not, so the evidence supplies at least $d_2(0.95\|q)$ bits. Its score is the budget gap, required minus supplied bits. We run the published adjudicator \citep{chlon2026pipeline} unchanged on VitaminC contrastive evidence \citep{schuster2021vitaminc} with Qwen2.5-7B-Instruct and GPT-4.1-mini as verifiers, holding each claim fixed while its evidence changes.

The 800 examples contain 395 \textsc{Supports}, 120 \textsc{Not Enough Info} (NEI) and 285 \textsc{Refutes} cases. Sufficiency AUROC compares \textsc{Supports} with NEI (515 examples); full-grounding AUROC treats NEI and \textsc{Refutes} together as unsupported (800 examples).

\begin{table}[t]
\caption{Evidence adjudication on VitaminC. Higher scores indicate less support. Flag counts are the adjudicator's fixed decisions at target 0.95. The 390 claim groups pair the same claim with supporting and with NEI or refuting evidence.}
\label{tab:nlp}
\centering\footnotesize
\begin{tabular}{@{}lrr@{}}
\toprule
Metric & Qwen2.5-7B & GPT-4.1-mini\\
\midrule
Sufficiency AUROC (515) & 0.9126 & 0.9143\\
Full-grounding AUROC (800) & 0.8990 & 0.9297\\
NEI flagged / 120 & 114 & 112\\
\textsc{Refutes} flagged / 285 & 261 & 275\\
\textsc{Supports} flagged / 395 & 103 & 99\\
Correct paired direction / 390 & 370 & 371\\
\bottomrule
\end{tabular}
\end{table}

\begin{figure}[t]
\centering
\includegraphics[width=0.6\textwidth]{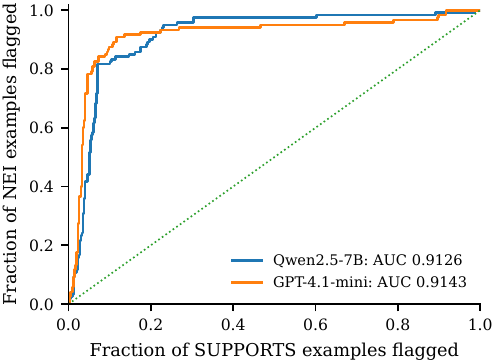}
\caption{Sufficiency ROC from the budget gap on 395 supporting and 120 NEI examples. Table~\ref{tab:nlp} adds the full-grounding result with all 285 refuting examples.}
\label{fig:roc}
\end{figure}

The budget gap separates supporting from insufficient evidence with AUROC 0.9126 for Qwen and 0.9143 for GPT-4.1-mini, and supporting from all unsupported evidence with AUROC 0.8990 and 0.9297 (Table~\ref{tab:nlp} and Figure~\ref{fig:roc}). For the same claim, supporting evidence receives the lower score in 370 and 371 of 390 claim groups. At its fixed operating point the adjudicator flags 95.0\% and 93.3\% of NEI cases and 91.6\% and 96.5\% of refuting cases, with flag rates of 26.1\% and 25.1\% on supporting evidence (Appendix~\ref{app:nlp}).

\subsubsection{Learning curves in context}\label{sec:countsfull}
Table~\ref{tab:counts} gives the curves of Figure~\ref{fig:counts}. In every one of the eight chair pairs, the eight patches that do not observe the feature leave the feature sites costlier than the anchor points alone, and the feature patch alone brings them to between 0.001 and 0.22 bits per site. One observing source view codes the depth better than one view drawn at random in all five scenes, and better than all four views together in scenes 4, 29 and 48. With three, four and five support cars, the order of the learner's own valuation codes better than the mean of the random orders in six, five and six of the eight cars.
\begin{table}[h]
\caption{Learning curves in context: mean held-out code length at each number of context items, by order of arrival (Figure~\ref{fig:counts}). Occupancy: bits per feature site, eight pairs, count 0 is the 300 anchor points alone. Stereo: bits per depth query, five scenes, 3,856 pixels. Kriging: log-density bits per pressure value, eight cars.}
\label{tab:counts}
\centering\small
\setlength{\tabcolsep}{5pt}
\begin{tabular}{@{}lrrrrrrrrrr@{}}
\toprule
Occupancy network, patches & 0 & 1 & 2 & 3 & 4 & 5 & 6 & 7 & 8 & 9\\
\midrule
Feature patch first & 4.99 & 0.046 & 0.021 & 0.019 & 0.014 & 0.016 & 0.014 & 0.015 & 0.015 & 0.015\\
Random order & 4.99 & 5.13 & 5.26 & 4.68 & 4.31 & 3.86 & 3.40 & 2.78 & 1.96 & 0.015\\
Feature patch last & 4.99 & 5.71 & 6.39 & 7.25 & 8.01 & 8.25 & 8.72 & 8.90 & 9.06 & 0.015\\
\midrule
Stereo network, source views & 1 & 2 & 3 & 4 &  &  &  &  &  & \\
\midrule
Observing views first & 8.71 & 8.61 & 9.73 & 10.17 &  &  &  &  &  & \\
All subsets of the size & 13.65 & 10.97 & 10.14 & 10.17 &  &  &  &  &  & \\
Non-observing views first & 15.70 & 11.04 & 10.81 & 10.17 &  &  &  &  &  & \\
\midrule
Kriging learner, support cars & 2 & 3 & 4 & 5 & 6 &  &  &  &  & \\
\midrule
Highest own valuation first & $-$0.43 & $-$1.61 & $-$1.78 & $-$1.88 & $-$1.95 &  &  &  &  & \\
Random order & 0.05 & $-$0.81 & $-$1.41 & $-$1.82 & $-$1.95 &  &  &  &  & \\
\bottomrule
\end{tabular}
\end{table}

\subsection{Choosing the channel on a trained transformer}\label{sec:decisionfull}
\begin{figure*}[t]
\centering
\includegraphics[width=\textwidth]{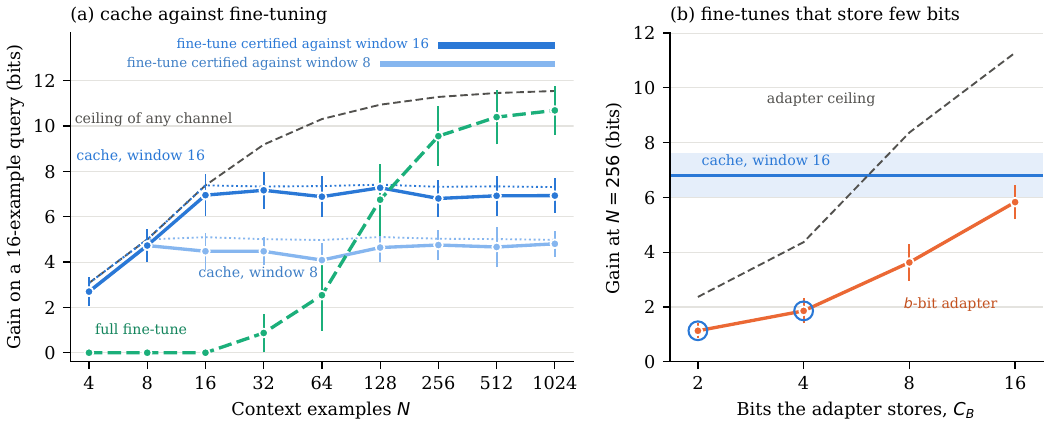}
\caption{Choosing between the cache and fine-tuning on a trained transformer, with exact information terms. (a) Gain on a 16-example query against the number of context examples. Dotted lines are the ceilings $I(Q;D_w)+\KL_0$ of the two caches and the dashed line is the ceiling $I(Q;D)+\KL_0$ of any channel; the bars mark the data sizes at which the fine-tune's gain passes the cache's ceiling, which certifies fine-tuning. (b) Gain of $b$-bit adapters at $N=256$ against the bits they store, with their ceilings $\min\{I(Q;D),C_B\}+\KL_0$ and the cache with a window of 16; rings mark the adapters against which the cache is certified. Intervals are 95\% over 64 tasks.}
\label{fig:decisionfull}
\end{figure*}

Proposition~\ref{prop:choose} is tested where the information terms are exact. A three-layer transformer with 104,194 parameters is trained to learn the tasks of Appendix~\ref{sec:closedforms} in context ($d=8$, $\rho=4$); each token carries the current input and the previous example, and the head returns a Gaussian predictive for the next label. Its code length is within 0.025 bits per label of the Bayes predictive on average. A query is a block of 16 examples whose labels are teacher-forced, and the context holds $N$ further examples of the same task, for $N$ from 4 to 1,024. Three channels share the trained weights:
\begin{itemize}
\item \textbf{the cache} reads the last $W$ context examples and then the query block, with a window of $W=8$ or $W=16$;
\item \textbf{the full fine-tune} trains all weights on the $N$ examples with Adam and then reads the query block alone, with its learning rate and number of steps chosen for each $N$ on separate tasks to maximise its gain;
\item \textbf{$b$-bit adapters} train $k$ pseudo-labels for fixed inputs, placed in front of every sequence, and quantise them to $b$ bits, so that they store at most $C_B=kb$ bits, for $(k,b)$ equal to $(2,1)$, $(4,1)$, $(8,1)$ and $(8,2)$.
\end{itemize}
Given the inputs, $I(Q;D)$ and $I(Q;D_w)$ are Gaussian log-determinants and $\ell^*(Q)$ is a Gaussian log-density. A certificate is issued when the lower end of the 95\% interval of its margin over 64 tasks is positive, and the measured winner is the channel whose paired gain difference has a 95\% interval that excludes zero (Appendix~\ref{app:decision}).

\begin{table}[t]
\caption{Certified and measured winners across data sizes: nine for the full fine-tune ($N$ from 4 to 1,024) and seven for the adapters ($N$ from 4 to 256). The columns `cache' and `tune' count the data sizes at which the cache certificate and the fine-tune certificate of Proposition~\ref{prop:choose} fire; `won' gives the data sizes won by the cache, won by the fine-tune and tied. For $N\le8$ both windows hold the whole context, so those comparisons appear under both; all 25 distinct certified calls agree with the measured winner.}
\label{tab:decision}
\centering\footnotesize
\setlength{\tabcolsep}{4pt}
\begin{tabular}{@{}lrcccccc@{}}
\toprule
 & & \multicolumn{3}{c}{Window 8} & \multicolumn{3}{c}{Window 16}\\
\cmidrule(lr){3-5}\cmidrule(l){6-8}
Fine-tune & $C_B$ & cache & tune & won & cache & tune & won\\
\midrule
Full & $\infty$ & 0 & 4 & 4/4/1 & 0 & 3 & 5/3/1\\
Adapter $2\times1$ & 2 & 6 & 0 & 7/0/0 & 6 & 0 & 7/0/0\\
Adapter $4\times1$ & 4 & 0 & 0 & 7/0/0 & 5 & 0 & 7/0/0\\
Adapter $8\times1$ & 8 & 0 & 0 & 6/0/1 & 0 & 0 & 7/0/0\\
Adapter $8\times2$ & 16 & 0 & 2 & 2/4/1 & 0 & 0 & 7/0/0\\
\bottomrule
\end{tabular}
\end{table}

\paragraph{Every certified call agrees with the measured winner.} Of the 64 distinct comparisons between a cache and a fine-tune (for $N\le8$ both windows hold the whole context and give the same cache), 25 are certified, and every certified call agrees with the measured winner on the same tasks (Table~\ref{tab:decision}). The cache's gain stops growing once the context passes the window, at 4.1--4.8 bits for $W=8$ and 6.8--7.3 bits for $W=16$, within 0.9 bits of its ceiling $I(Q;D_w)+\KL_0$ (Figure~\ref{fig:decisionfull}a). The fine-tune's gain keeps rising with the data: it rises from 0.9 bits at $N=32$ to 10.7 bits at $N=1{,}024$, against a ceiling of 11.5 bits for any channel. It beats the cache from $N=128$ with the window of 8 and from $N=256$ with the window of 16, and at those data sizes and every larger one its gain passes the cache's ceiling, which certifies fine-tuning against every cache with that window. In both cases the full fine-tune first wins at sixteen times the window, after a tie at eight times: the larger window leaves less evicted information for fine-tuning to exploit.

\paragraph{When the context fits.} For $N\le W$, every fine-tuning schedule lowered the gain on the tuning tasks, so the tuned recipe leaves the weights unchanged, and the cache wins with 2.7 to 7.0 bits. Part (iii) caps what any fine-tune could have added there at the cache's decoding loss, estimated at 0.27 to 0.43 bits.

\paragraph{A fine-tune that stores few bits never catches up.} Each adapter's gain stays below its ceiling $\min\{I(Q;D),C_B\}+\KL_0$: at $N=256$ the adapters storing 2, 4, 8 and 16 bits gain 1.1, 1.9, 3.6 and 5.8 bits (Figure~\ref{fig:decisionfull}b). The cache with the window of 16 saves 6.8--7.3 bits, more than the 2-bit and 4-bit adapters can store with the prior repair of 0.37 bits added, so part (iv) says they lose at every $N\ge16$; they lose at every data size tested, and the certificate fires at every $N\ge16$ for both. The 16-bit adapter, whose ceiling exceeds the cache's gain, beats the cache with the window of 8 from $N=32$; its gain exceeds that cache's ceiling from $N=32$ on, and the fine-tune certificate fires at $N=64$ and $N=256$.

\subsection{The choice across domains in full}\label{app:domainsfull}
Table~\ref{tab:domainruns} lists every run behind Section~\ref{sec:domains}. On both tables TabPFN encodes the held-out labels in fewer bits than XGBoost in every seed, and the test calls it in every seed and in all 200 orders. On airfoil pressure the trained surrogate encodes the fields in fewer bits in every seed, and the test calls it after 53 to 66 queries. The original pair of airfoil predictors, fitted with 149, 149 and 145 calibration designs before the fresh cohort was drawn, is called the same way after 70 to 122 queries. Table~\ref{tab:curves} gives the learning curves of Figure~\ref{fig:domains}a with the call at each size, and Table~\ref{tab:controls} the simulated streams of Figure~\ref{fig:domains}c.
\begin{table}[h]
\caption{The anytime-valid choice across domains, run by run. Bits are the mean held-out code length per label (tables) or per pressure value (airfoil) without mixing; accuracy is for the tables only. Call and queries are for each seed's designated query order; the last column counts the 200 random orders of the same cohort that reach the same call.}
\label{tab:domainruns}
\centering\small
\begin{tabular}{@{}llrrrlrr@{}}
\toprule
Comparison & Seed & In context & Trained & Accuracy (\%) & Call & Queries & Orders\\
\midrule
Phoneme: TabPFN vs XGBoost & 0 & 0.587 & 0.761 & 78.2 / 78.3 & in context & 348 & 200/200\\
 & 1 & 0.605 & 0.691 & 76.6 / 78.9 & in context & 1,390 & 200/200\\
 & 2 & 0.640 & 0.965 & 77.9 / 78.2 & in context & 151 & 200/200\\
\addlinespace[2pt]
Spambase: TabPFN vs XGBoost & 0 & 0.379 & 0.716 & 91.5 / 85.8 & in context & 198 & 200/200\\
 & 1 & 0.299 & 0.501 & 92.7 / 88.7 & in context & 247 & 200/200\\
 & 2 & 0.318 & 0.454 & 91.7 / 88.1 & in context & 358 & 200/200\\
\addlinespace[2pt]
Airfoil: support set vs MLP & 0 & 3.654 & 2.412 & --- & trained & 62 & 200/200\\
 & 1 & 3.623 & 2.384 & --- & trained & 66 & 200/200\\
 & 2 & 3.670 & 2.227 & --- & trained & 53 & 200/200\\
\addlinespace[2pt]
Airfoil, original predictors & 0 & 3.787 & 3.063 & --- & trained & 122 & 200/200\\
 & 1 & 3.654 & 2.399 & --- & trained & 71 & 200/200\\
 & 2 & 3.705 & 2.566 & --- & trained & 70 & 200/200\\
\bottomrule
\end{tabular}
\end{table}

\begin{table}[h]
\caption{Airfoil pressure against the number of designs in the support set (in context) or the training set (trained), with the calibration designs and the 254 fresh queries fixed in each seed. Bits are the mean held-out code length per pressure value over three seeds; the last column gives the anytime-valid call in each seed's designated order (seeds 0, 1, 2). The trained surrogate's recipe stops early on a validation split, which needs at least 16 designs.}
\label{tab:curves}
\centering\small
\begin{tabular}{@{}rrrl@{}}
\toprule
Designs & In context & Trained & Call in seeds 0, 1, 2\\
\midrule
4 & 4.964 & --- & ---\\
8 & 4.718 & --- & ---\\
16 & 4.583 & 4.746 & none, context, none\\
32 & 4.258 & 4.292 & trained, context, context\\
64 & 3.951 & 3.573 & trained, context, trained\\
128 & 3.649 & 2.341 & trained, trained, trained\\
256 & 3.458 & 2.382 & trained, trained, trained\\
512 & 3.298 & 1.516 & trained, trained, trained\\
\bottomrule
\end{tabular}
\end{table}

\begin{table}[h]
\caption{The anytime-valid test on 10,000 simulated streams per case, each of up to 4,096 queries, with the settings of the real comparisons and the bound of the table comparisons, $c=\log_2(2/\varepsilon)$. Advantage is the mean score difference in bits per query, with independent $\pm2$-bit noise for the nonzero cases; the nulls have mean zero. Wrong calls are calls for a side that is not better, with the upper end of a 95\% Wilson interval; correct calls are shares of all streams by the given number of queries.}
\label{tab:controls}
\centering\small
\begin{tabular}{@{}lrrrrr@{}}
\toprule
Case & Wrong calls & 95\% upper & \multicolumn{3}{c}{Correct calls by 256 / 1,024 / 4,096}\\
\midrule
Exact tie & 0.00\% & 0.04\% & \multicolumn{3}{c}{---}\\
Symmetric null & 3.25\% & 3.62\% & \multicolumn{3}{c}{---}\\
Skewed null & 2.87\% & 3.22\% & \multicolumn{3}{c}{---}\\
Null of changing variance & 3.31\% & 3.68\% & \multicolumn{3}{c}{---}\\
Advantage 0.5 & 0.00\% & 0.04\% & 88.1\% & 100.0\% & 100.0\%\\
Advantage 0.1 & 0.17\% & 0.27\% & 3.7\% & 12.3\% & 54.9\%\\
Advantage 0.03 & 0.74\% & 0.93\% & 1.1\% & 2.2\% & 4.5\%\\
Advantage 0.005 & 1.21\% & 1.44\% & 0.9\% & 1.3\% & 1.8\%\\
Advantage $-0.1$ & 0.16\% & 0.26\% & 4.0\% & 12.3\% & 54.7\%\\
\bottomrule
\end{tabular}
\end{table}

\subsection{The measurable tools on GPT-2}\label{sec:gpt2full}
\begin{figure*}[t]
\centering
\includegraphics[width=\textwidth]{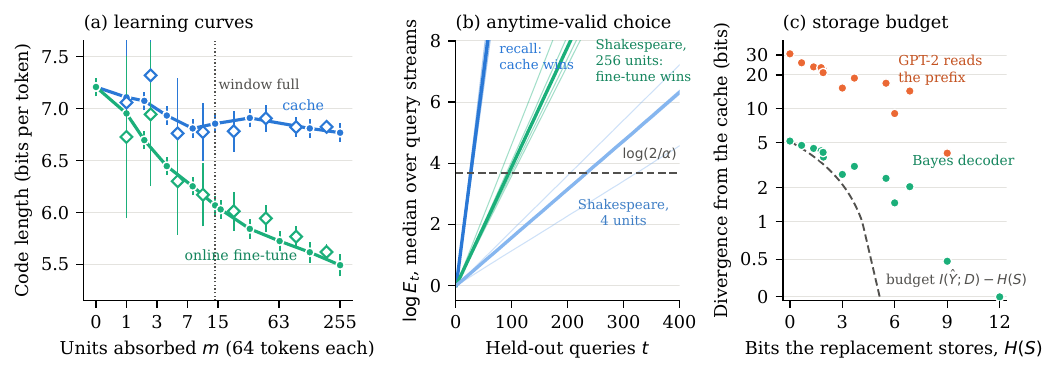}
\caption{The measurable tools of Section~\ref{sec:choose} on GPT-2. (a) Learning curves of the cache and the online fine-tune on Shakespeare passages: code length on held-out passages (circles) and the prequential increments read from the context itself, binned over sizes (diamonds); the dotted line marks the full window. Intervals are $\pm1.96$ standard errors of the per-unit means, pooled over four streams. (b) $\log E_t$ of the test that calls the winner, median over 2,000 query streams for each context (thin lines) and the median over contexts (thick lines): recall of 512-token passages against a one-epoch fine-tune, where the cache wins, and Shakespeare with 256 and 4 context units, where the fine-tune wins; the dashed line is the threshold $\log(2/\alpha)$. (c) Total divergence from the cache over the four-key workload against the bits a compiled replacement stores, for the replacement's Bayes decoder and for GPT-2 reading it; the dashed line is the budget of Theorem~\ref{thm:storage}.}
\label{fig:gpt2}
\end{figure*}

Section~\ref{sec:choose} reads both channels without the true distribution. We run its three measurable tools on GPT-2, the pretrained 124M-parameter model of \citet{radford2019gpt2}, with the original weights \citep{onnxmodelzoo} (Appendix~\ref{app:gpt2}).

\paragraph{Learning curves from the context.}
The units are 64-token passages of Shakespeare \citep{karpathy2015charrnn}, split at random into a context pool and a held-out pool. The cache is GPT-2 reading the previous units as separate passages in its 1,024-token window, which holds the last 14; the online fine-tune takes one Adam step on each unit as it arrives. Over four streams of 256 units, the cache's increments are read by teacher-forcing each unit after the units in its window, and the fine-tune's are its losses before each step (Proposition~\ref{prop:preq}). In all eight size bins and for both channels, the increments agree with the held-out learning curves within the sampling error of the per-unit means, at most 1.97 standard errors apart; from 14 units on they lie within 0.16 bits per token of the curves. Their sums, the prequential code lengths, come within 0.1\% for the cache and 1.5\% for the fine-tune of the sums of the held-out curves (Figure~\ref{fig:gpt2}a). Both readings make the same call in every bin. The cache levels off at 6.8 to 6.9 bits per token from 8 units on, while the fine-tune keeps improving: on the mean curves it leads at every size, by 0.16 bits per token after one unit and by 1.28 after 256, where it reaches 5.49.

\begin{table*}[t]
\caption{The anytime-valid choice on GPT-2. Code lengths are in bits per token on each task's pool of held-out queries, ranges over the four contexts (streams or passages). `Calls' counts the query streams, 2,000 per context, in which the test called the better memory, and `queries' gives the median number of held-out queries to the call over the streams that called, range over contexts. In the 4,096-token, one-epoch row the fine-tune is better on three passages and the cache on one; there 2,048 streams are still undecided after 1,000 queries and 4 call the losing memory. No other stream calls the losing memory.}
\label{tab:gpt2}
\centering\footnotesize
\begin{tabular}{@{}lllcclrr@{}}
\toprule
 & & & \multicolumn{2}{c}{Code length} & & & \\
\cmidrule(lr){4-5}
Task & Context & Recipe & Cache & Fine-tune & Better memory & Calls & Queries\\
\midrule
Shakespeare & 4 units & online & 6.98--7.23 & 6.59--6.68 & fine-tune & 8,000/8,000 & 205--324\\
Shakespeare & 256 units & online & 6.80--7.13 & 5.68--5.70 & fine-tune & 8,000/8,000 & 81--103\\
Recall & 512 tokens & 1 epoch & 0.62--1.92 & 5.44--6.64 & cache & 8,000/8,000 & 25--29\\
Recall & 512 tokens & 4 epochs & 0.62--1.92 & 2.91--4.46 & cache & 8,000/8,000 & 46--69\\
Recall & 512 tokens & 16 epochs & 0.62--1.92 & 0.23--0.36 & fine-tune & 8,000/8,000 & 70--300\\
Recall & 4,096 tokens & 1 epoch & 5.16--5.52 & 4.70--5.32 & fine-tune (3), cache (1) & 5,948/8,000 & 165--668\\
Recall & 4,096 tokens & 4 epochs & 5.16--5.52 & 2.44--2.96 & fine-tune & 8,000/8,000 & 41--53\\
Recall & 4,096 tokens & 16 epochs & 5.16--5.52 & 0.30--0.39 & fine-tune & 8,000/8,000 & 25--26\\
\bottomrule
\end{tabular}
\end{table*}

\paragraph{An anytime-valid choice.}
The test of Theorem~\ref{thm:eprocess} runs in both directions on one stream of held-out queries, each direction at level 0.025, and the first to reach 40 makes the call (Appendix~\ref{app:gpt2}). On Shakespeare the online fine-tune is the better memory with 4 context units and with 256, and the test calls it in all 16,000 query streams, after a median of 205--324 and 81--103 queries (Table~\ref{tab:gpt2}). In document recall a query is a span of a passage, cued by its first 8 tokens and scored on the next 24, and GPT-2 copies the passage from its window. On passages of 512 tokens the cache needs 0.62--1.92 bits per token, against 2.91--6.64 for fine-tunes of one and four epochs, and the test calls the cache after a median of 25--69 queries; a fine-tune of sixteen epochs memorises the passage, needs 0.23--0.36 bits per token, and the test calls it after 70--300. On passages of 4,096 tokens the window holds the last 960, and fine-tunes of four and sixteen epochs win and are called after 25--53 queries. After one epoch the fine-tune and the cache lie within 0.76 bits per token of each other on these passages: the test calls the better memory in 5,948 of 8,000 streams and leaves 2,048 undecided after 1,000 queries, 2,046 of them in the two comparisons separated by 0.17 and 0.21 bits per token. Over all 64,000 query streams, 4 call the losing memory, all in the comparison separated by 0.21 bits per token and within the 2.5\% error the test allows there, and on increments centred at zero the procedure reaches its threshold in at most 0.66\% of 10,000 streams for any of the 32 comparisons (Figure~\ref{fig:gpt2}b).

\paragraph{A lower bound on GPT-2's decoding loss.}
Four units fit the window, and the cache keeps them (Proposition~\ref{prop:capacity}), so by Proposition~\ref{prop:choose}(iii), applied to each context as in Appendix~\ref{app:choose}, the online fine-tune's advantage over the cache on each context is a lower bound on the cache's decoding loss there. In each of the four streams the fine-tune beats the cache by 0.35 to 0.55 bits per token on held-out passages, with standard errors below 0.03: on each of these contexts GPT-2's prediction falls at least that far short of the best prediction from its prompt. The held-out passages are drawn independently of those in context, so beyond the task the context carries no information about them, and the shortfall is prior repair that the cache leaves unrealised.

\paragraph{The storage budget.}
The context is a table of four facts, each a fixed key of three tokens followed by one of eight candidate tokens, so it ranges over $8^4=4{,}096$ tables and holds 12 bits. The workload is the four keys, and an output is GPT-2's next token after a key, read as one of the candidates or as none of them. From GPT-2's own in-context outputs over all 4,096 tables, $I(\hat Y_{1:4};D)=5.14$ bits: the cache carries 43\% of the table into its outputs. Equation~\eqref{eq:Ihat} with $M=8$, 32, 128 and 512 tables drawn at random gives 2.57, 3.75, 4.52 and 4.93 bits, each below $\log_2M$, and 5.14 bits with all 4,096. A compiled replacement of the cache lists the first $k$ facts with their words restricted to the first $2^b$ candidates, and each table takes the state that GPT-2 reads closest to its cache, so the replacement stores $H(S)$ bits about the table. The 13 replacements store from 0 to 12 bits, and the eight that store less than 5.14 bits all miss the cache by at least $5.14-H(S)$ bits on the workload (Figure~\ref{fig:gpt2}c). The divergence splits as in the proof of Theorem~\ref{thm:storage}. With the Bayes decoder of the state, which leaves only the storage term, the replacement that stores nothing misses by 5.18 bits against a budget of 5.14, and the three that list one fact stay within 0.52 bits of the budget. GPT-2 reading the same compiled prefix adds 3.6 to 25.6 bits of decoding loss to every replacement short of the full table.

\subsection{Certificates on pretrained models outside language}\label{app:certnative}
Proposition~\ref{prop:choose} is tested on three pretrained models outside language, each on a designed distribution of contexts for which the information terms are exact. For any channel $c$, $\E[\Gamma_c]-\KL_0=H(Q)-\E[\ell_c]$, so the prior repair cancels from both certificates of part (ii): a memory beats every context memory confined to the window $D_w$ when its code length falls below $H(Q\mid D_w)$, and the in-context model beats every memory that stores at most $C$ bits about the context, any fine-tune, adapter or compiled state of that size, when its code length falls below $H(Q)-C$. Each model runs 64 episodes, every predictor is mixed with $\varepsilon=0.01$ of the uniform law, a certificate is issued when the 95\% interval of its paired margin lies above zero, and each issued certificate is checked against the measured winner of every implemented pair it covers, as in Appendix~\ref{sec:decisionfull}. Of 96 certificates tested, 36 are issued, and all 86 checks agree with the measured winner (Table~\ref{tab:certnative}).

\begin{table}[h]
\caption{Certificates on pretrained models outside language, 64 episodes per model: certificates issued out of those tested, against a window and against a capacity, and checks of the issued certificates against the measured winners of the implemented pairs they cover.}\label{tab:certnative}
\centering\small
\begin{tabular}{@{}lrrrrr@{}}
\toprule
Model & Window & Capacity & Checks & Agree & Disagree\\
\midrule
TabPFN-v2 & 8/30 & 17/48 & 50 & 50 & 0\\
Occupancy network & 2/5 & 3/6 & 14 & 14 & 0\\
MVSNet & 2/3 & 4/4 & 22 & 22 & 0\\
\midrule
All & 12/38 & 24/58 & 86 & 86 & 0\\
\bottomrule
\end{tabular}
\end{table}

\paragraph{TabPFN.} Tasks are drawn uniformly from a pool of 4,096 logistic classifiers on four Gaussian features, with weights of standard deviation 3 and biases of standard deviation 1. A context holds $N=8$ to 1,024 labelled rows of one task, nested across sizes, and a query is a block of 64 further rows. The pool makes the true posterior predictive a finite mixture, so $H(Q)$, $H(Q\mid D)$ and $H(Q\mid D_w)$ are exact in each episode, and $I(Q;D)$ rises from 3.4 bits at $N=8$ to 9.5 bits at $N=1{,}024$. TabPFN-v2 reads all rows in its context, or only the last 16 or 64; logistic regression and XGBoost train on all rows; and the bounded memories are logistic regressions whose weights and bias lie on a grid of $2^b$ values, fitted over the whole grid, which store at most $5b$ bits for $b=1,2$, and the best of $2^k$ fixed classifiers by likelihood, which stores at most $k$ bits for $k=2$ to 10. With 1,024 rows TabPFN codes a block within 0.67 bits of the true posterior predictive. The certificate for learning in context fires from $N=64$ on, and at $N=1{,}024$ it covers every memory of at most 8 bits. Logistic regression, and TabPFN reading all rows, are certified against every memory confined to the last 16 rows from $N=128$ on. Against the last 64 rows, whose ceiling the trained models' code lengths reach at 512 and 1,024 rows, no certificate is issued.

\paragraph{Occupancy network.} A procedural chair carries six bumps, each present or absent with probability $1/2$. The context is a point cloud read as items in a random order: 300 anchor points on the shared geometry, always read; one patch of 300 points around each bump, which lies on the bump if it is present and on the flat surface beneath it if not; and six patches of 300 points from the shared geometry. The query is the occupancy of one site inside each bump, averaged over 32 sites per bump, so $H(Q)=6$ bits, $H(Q\mid D)=0$, and $H(Q\mid D_w)$ counts the bumps whose patch lies outside the first $W$ items. The pretrained occupancy network, a three-plane model for partial point clouds, reads every item and codes the six sites in 2.72 bits, so it is certified against every memory that stores at most 2 bits about the shape, and against every memory confined to the first 2 or 4 items, whose ceilings are 5.00 and 3.97 bits. Reading its anchors alone, the same network codes the sites in 19.4 bits against a ceiling of 6: where no patch observes a bump it predicts with confidence, and the excess is decoding loss.

\paragraph{Stereo network.} A textured card faces the reference camera at one of 16 depths, drawn uniformly, in front of a textured background at a fixed depth. Textures are defined in reference-image coordinates, so the reference image is the same at every depth. One of four source views observes the card; the other three are turned away so that the card lies outside their frame at every designed depth, and the views are read in a random order. The query is which of the 16 depths the card lies at, read from the network's law over its 192 hypotheses at pixels inside the card, so $H(Q)=4$ bits, $H(Q\mid D)=0$, and $H(Q\mid D_w)=4$ bits unless the observing view is among the first $W$ read. MVSNet with the checkpoint of Appendix~\ref{app:mvs}, reading all views, codes the depth in 0.67 bits, so it is certified against every memory that stores at most 3 bits about the scene, and against every memory confined to the first one or two views.

\paragraph{Bounded memories of the two networks.} The certificate for learning in context covers every memory of at most $C$ bits. For the two networks it is checked against compiled states, which store $k$ of the bumps or the top $k$ bits of the depth index and are read by their Bayes decoder and by the frozen network, given the stored bumps' patches or a depth range narrowed to the stored bits.

\section{Experimental protocols}\label{app:protocols}
\subsection{Geometry: local features and natural objects}\label{app:geometry}
\paragraph{Controlled feature construction.}
The preparation seed is 20261017. Eight independently seeded parametric chair pairs differ by a raised seat feature versus a recess of matched extent, with unchanged surrounding geometry and a common frame. Sampling starts from 100,000 area-uniform surface samples, applies the partial-crop policy, draws 3,000 points with replacement and adds Gaussian noise of standard deviation 0.005. Sampling allocations and random draws are paired wherever the geometry is shared, and the deletion units are built before the model is evaluated.

A pair is eligible when both distinguishing surfaces survive the crop, the changed observations fit within the 300-point item, and at least 300 anchor observations remain outside the feature plus a 0.01 safety band; eligibility is decided from geometry alone. Three site populations are fixed from geometry: 1,024 feature-positive sites, 1,024 unchanged occupied sites and 1,024 unchanged free sites. The main text reports the feature-positive population, and all three are in \texttt{geometry\_feature.csv}. The mesh ambiguity tolerance is $10^{-5}$.

The relevant deletion removes the distinguishing observations, while the coherent absent arm supplies the observed recess floor and walls. Under the relevant deletion the paired remaining inputs agree exactly. The pretrained model is used without fine-tuning.

\paragraph{Donor replacement.}
The cohort contains two objects in each of eight ShapeNet categories. Donor replacement keeps 300 target anchor points and substitutes the optional observations of another object in the same category. The 8,492 claims are those that the target's optional observations lift above 0.95, defined from the target's own predictions, so donor confidence never enters their selection. Donor-mesh agreement uses the same geometric predicate as the claim. Table~\ref{tab:donor} gives pooled and equal-object rates.

\paragraph{Coherent-object replacement.}
This protocol replaces the entire observation set, anchors included. Query strata (target and donor interior disagreement, exterior, and near-boundary) are sampled before prediction, with a boundary band of 0.01 in the normalized frame and query locations in $[-0.55,0.55]^3$. Mesh occupancy uses validated distance checks and paired ray-parity checks, with boundary ambiguity as a separate state. The protocol requests 1,024 queries per stratum, subject to geometric eligibility, and measures occupancy at fixed coordinates where the two shapes disagree. Each target is compared with every eligible donor among the other 15 objects (237 comparisons).

\begin{table}[h]
\caption{Donor replacement by the donor mesh's relation to the fixed claim. Equal-object rates differ from pooled rates because objects contribute different numbers of claims.}
\label{tab:donor}
\centering\small
\begin{tabular}{@{}lrrrr@{}}
\toprule
Donor mesh relation & Claims & Still at 0.95 & Pooled rate & Equal-object rate\\
\midrule
Agrees with fixed claim & 5,902 & 5,305 & 89.88\% & 78.34\%\\
Disagrees with fixed claim & 2,589 & 64 & 2.47\% & 2.30\%\\
Boundary ambiguous & 1 & 0 & 0\% & 0\%\\
\bottomrule
\end{tabular}
\end{table}

\subsection{Multi-view preprocessing, claims, and visibility}\label{app:mvs}
The native 3D-CNN MVSNet checkpoint is \texttt{model.ckpt-150000}. Preprocessing uses maximum width 1,152 and height 864 with aspect-preserving scaling, cropping and matching camera updates, and the memory-efficient native inference path exposes the categorical law on 192 depth hypotheses. The reference image and camera are present in every condition, the four source views follow the original pairing order, and the physical depth hypotheses are identical across deletions.

Scene 9 was used to develop the protocol. Scenes 4, 48, 11 and 29 were drawn with seed 20261017 from the MVSNet test scenes. The first five eligible reference views per scene and their first four distinct ranked source views are fixed without reference to predictions, and each reference starts from 4,096 seeded pixel queries sampled after dense inference.

The depth claim is an interval centred at the full-context expected depth with half-width five depth units. A reference depth more than one unit inside the interval defines the oracle-inside stratum, more than one unit outside defines oracle-outside, and the remaining cases form a boundary stratum; invalid reference depths are counted separately. The same intervals are used under every deletion.

Visibility projects the reference surface point into each source and compares the independently provided source depth, with absolute tolerance one depth unit, relative tolerance 0.002 and reprojection tolerance one original-image pixel. The categories are observing, occluded, outside or behind, and unknown. The observing and non-observing deletions take the lowest original slot index within their category, and unknown views are never treated as non-observing. The paired oracle-inside cohort contains 11,808 query locations across 25 references.

Means are computed within reference and stratum, then over references within scene, then over scenes, and crossing counts use the same paired cohorts. Table~\ref{tab:scenes} reports every scene-level contrast, and the bootstrap resamples scenes. The evaluated pipeline matches the native MVSNet implementation exactly in 12 preprocessing, forward-pass and condition comparisons and in two further regression comparisons on scene 4.

\begin{table}[h]
\caption{Extra claim-probability loss from deleting an observing rather than a non-observing view, in percentage points. Five references contribute equally within each scene.}
\label{tab:scenes}
\centering\small
\begin{tabular}{@{}lrrr@{}}
\toprule
Scene & Oracle inside & Oracle outside & Boundary\\
\midrule
9 (protocol development) & 3.3232 & 0.7735 & 1.9261\\
4 & 3.9743 & $-0.7445$ & 6.3977\\
48 & 13.8833 & 4.1997 & 7.1883\\
11 & 10.6232 & 3.7646 & 4.4055\\
29 & 15.7790 & 2.5745 & 11.3640\\
\midrule
Mean, all five scenes & 9.5166 & 2.1136 & 6.2563\\
Mean, scenes 4, 48, 11 and 29 & 11.0650 & 2.4486 & 7.3389\\
\bottomrule
\end{tabular}
\end{table}

\subsection{Kriging learner and native inference}\label{app:fields}
The learner receives each support car as points, normals, positive area weights and surface pressure coefficients; the query supplies geometry only. Chart anchors are farthest points of the support union, and local descriptors contain kernel-mean offsets and normals. Centred Gram systems give affine kriging weights summing to one, followed by chart smoothing, and kernel transport uses position and normal agreement to carry support pressures to the query. Descriptor scale, transport scale, ridge and smoothing are selected by leave-one-car-out prediction on the support set, so the learner is a fixed inference procedure with context-dependent fitted state.

The spread transports each support car's leave-one-out squared-residual field to the query and averages these fields, with a floor of 0.05 times the median squared leave-one-out error and a small positive safeguard. The result is an explicit error-based Gaussian predictive law; any Gaussian predictive law, including a conventional GP posterior with fixed hyperparameters, enters the evaluator in the same way.

Each query has 32,000 prepared geometry points and 512 fixed scoring sites. Query geometry enters the local descriptors and the \texttt{coarsest} argument that sets descriptor bandwidth, so subsampling the geometry before inference changes the predictions even at the scored sites. The fidelity grid crosses two geometry sizes (512 and 32,000), two chart-anchor counts (128 and 1,024) and two leave-one-out selection sizes (512 and 4,000), and two further cells change the supplied geometry and the bandwidth argument separately. Support names and family order are fixed because the learner's selection rule uses name-seeded sampling.

Two claim banks are fixed in advance and held across every inference setting and deletion: the native-answer bank fixes the native full-context mean at each site, and the reduced-input bank fixes the reduced-input mean. Both use pressure tolerance 0.05 and target 0.95. Under native inference, the native-answer claim has full-context probability $2\Phi(0.05/s)-1$, where $s$ is the learner's predictive spread at the site. The gain does not depend on the bank, while claim-level bits and crossings do (Table~\ref{tab:fieldbanks}).

\begin{table}[h]
\caption{Field scores for both claim banks. Means weight sites by area within a query and weight queries and single-car deletions equally. Crossing fractions are percentages of area-weighted site observations.}
\label{tab:fieldbanks}
\centering\small
\begin{tabular}{@{}llrrrrrr@{}}
\toprule
Inference & Claim bank & $p$ & $q$ & Gain $I_G$ & Claim bits $J$ & Crossing & Reverse\\
\midrule
Reduced input & Reduced-input answer & 0.5761 & 0.5312 & 0.7939 & 0.1108 & 4.6045\% & 1.6273\%\\
Reduced input & Native answer & 0.4623 & 0.4806 & 0.7939 & 0.1056 & 1.4881\% & 2.6273\%\\
Native & Reduced-input answer & 0.4629 & 0.4584 & 0.2296 & 0.0382 & 1.2688\% & 1.3699\%\\
Native & Native answer & 0.6131 & 0.5895 & 0.2296 & 0.0419 & 4.2862\% & 2.2225\%\\
\bottomrule
\end{tabular}
\end{table}

Reference pressure enters only the error,
\begin{equation}\label{eq:relL2}
\text{relative }L_2=\sqrt{\frac{\sum_j w_j(\hat y_j-y^{\mathrm{ref}}_j)^2}{\sum_j w_j(y^{\mathrm{ref}}_j)^2}}.
\end{equation}
The reported mean is an equal-query mean of this ratio, and the query car is the unit of replication. Native inference improves all eight cars. The evaluated learner reproduces the original implementation's means, variances and selected configurations exactly in 168 comparisons. The learner needs at least two support cars, so every deletion leaves at least two.

\subsection{Language adjudication}\label{app:nlp}
The adjudicator is the Trace Budget implementation \citep{chlon2026pipeline}, run with Qwen2.5-7B-Instruct and with GPT-4.1-mini (version 2025-04-14). Temperature is zero, the top-label log-probability request is five, and the target is 0.95. Claim normalization, budget-gap extraction and the decision policy run unchanged, and the flag is the adjudicator's own decision rather than a threshold fitted to these data.

The score is the budget gap in bits, required minus supplied, so larger values indicate less support. The adjudicator clips probabilities to $[10^{-12},1-10^{-12}]$, which bounds the gap at $d_2(0.95\|10^{-12})=37.58$ bits; 47.0\% of Qwen scores and 51.5\% of GPT-4.1-mini scores lie within 0.1 bits of this bound, and AUROC uses the full score ordering with ties counted as half a pair.

The evaluation set is the first 800 VitaminC examples selected by the benchmark pipeline: 395 \textsc{Supports}, 120 NEI and 285 \textsc{Refutes}. The 390 claim groups contain the same claim with supporting evidence and with NEI or refuting evidence; the Qwen score orders 370 of them correctly with no ties, and the GPT-4.1-mini score orders 371 correctly with one tie.

Three GPT-4.1-mini requests (examples 762--764; two NEI and one \textsc{Supports}) returned content-filter errors and count as flagged. On the 797 successful examples, sufficiency AUROC is 0.9152 (394 \textsc{Supports} and 118 NEI) and full-grounding AUROC is 0.9318.

\subsection{Decision test}\label{app:decision}
\paragraph{In-context learner.} Tasks follow Appendix~\ref{sec:closedforms} with $d=8$, $\bar\tau^2=4$ and $\sigma^2=1$. The learner is a pre-norm causal transformer with three layers, width 64, four heads, MLP width 128 and learned positions, 104,194 parameters in all. Token $i$ is $[x_i,x_{i-1},y_{i-1},\one\{i>1\}]$, and the head at position $i$ returns the mean and log-variance of a Gaussian predictive for $y_i$, so position $i$ sees the $i-1$ preceding examples. Training draws 20,000 batches of 64 sequences of 32 examples from the prior and minimises the prequential Gaussian negative log-likelihood with AdamW (peak learning rate $10^{-3}$, 1,000 warm-up steps, cosine decay, weight decay $10^{-4}$, gradient clipping at 1). On 256 validation sequences its code length exceeds that of the Bayes posterior predictive by 0.025 bits per label on average.

\paragraph{Channels.} The base reads the 16-example query block alone. The cache reads the last $\min\{N,W\}$ context examples and then the block, and only the block's labels are scored. The full fine-tune trains all weights with Adam on $\ell(D)$, computed at every step on a fresh random split of the context into sequences of at most 16 examples. Its learning rate and number of steps are chosen for each $N$ to maximise the mean gain on 32 tuning tasks, from learning rates $10^{-4}$, $3\times10^{-4}$, $10^{-3}$, $3\times10^{-3}$ and $10^{-2}$ (with $3\times10^{-5}$ added for $N\in\{16,32\}$ and $3\times10^{-2}$ for $N\ge32$) and from 0, 5, 10, 20, 40 and 80 steps (up to 160 at $10^{-4}$ and 320 at $3\times10^{-5}$). Every chosen learning rate lies inside this grid. The chosen recipes leave the weights unchanged for $N\le16$, take 80 steps at $10^{-4}$ for $N=32$ and 40 steps at $10^{-2}$ for $N=64$, and take 40 steps at $3\times10^{-3}$ beyond. The adapters place $k$ pseudo-examples, with inputs fixed by a seed and trainable labels, in front of every sequence; the labels are trained with Adam (learning rate 0.1, 100 steps) on $\ell(D)$ and then quantised to $b$ bits on a uniform grid over $[-R,R]$, with $R$ chosen for each $N$ on the tuning tasks from $\{1,2,3,4\}$, or from $\{1,2,3,4,5,6,8\}$ for the 16-bit adapter. The quantised labels are the adapter's only trainable state, so it takes at most $2^{kb}$ values and $I(D;Z_B)\le kb$.

\paragraph{Evaluation.} The 64 evaluation tasks are shared by every channel and data size; each has a stream of 1,024 context examples and a 16-example query block, and the context of size $N$ is the first $N$ examples of the stream. Given the inputs, $I(Q;D)$ and $I(Q;D_w)$ are Gaussian log-determinants, $\ell^*(Q)$ is a Gaussian log-density, and the prior repair is the mean of $\ell_0(Q)-\ell^*(Q)$, 0.37 bits (95\% interval 0.12 to 0.61). The cache certificate uses the per-task margin $\ell^*(Q)-\ell_A(Q\mid D)-\min\{I(Q;D),C_B\}$ and the fine-tune certificate uses $\ell^*(Q)-\ell_B(Q)-I(Q;D_w)$; a certificate is issued when the lower end of the 95\% interval of the mean margin is positive. The measured winner is the channel whose paired gain difference has a 95\% interval that excludes zero, and the comparison is a tie otherwise. Adapters are run for $N\le256$, where their gains have already levelled off.

\subsection{GPT-2 experiments}\label{app:gpt2}
\paragraph{Model.} GPT-2 has 124,439,808 parameters; we use the OpenAI weights exported in the ONNX Model Zoo and their byte-level BPE vocabulary. The implementation reproduces GPT-2's behaviour: 3.99 bits per token on a Wikipedia paragraph, and 0.27 bits per token on the second copy of a repeated sequence of 50 random tokens, after its first token. Fine-tuning updates all weights with Adam.

\paragraph{Shakespeare streams.} The text is cut into 5,281 units of 64 tokens and permuted with a fixed seed: 1,024 units form the held-out pool and the rest the context pool. Each of four streams draws 256 context units in random order and 64 held-out units for its learning curves, from a sub-pool of 384; a further 256 held-out units define the task of the anytime-valid test. Both channels read every unit as a separate passage, the end-of-text token followed by its 64 tokens. The cache's context is the end-of-text token followed by the last 14 units, each closed by an end-of-text token, so the context and a query passage fit the 1,024-token window. The online fine-tune takes one Adam step per unit at learning rate $3\times10^{-5}$, chosen from $10^{-5}$, $3\times10^{-5}$ and $10^{-4}$ by the held-out code length after 64 units of a separate ordering of the context pool, on held-out units disjoint from those reported. Increments are read before each step for the fine-tune and with the window rule for the cache. Held-out learning curves are read at 9 sizes for the cache and 11 for the fine-tune; Figure~\ref{fig:gpt2}a bins the increments over $m$ in $\{1\}$, $\{2,3\}$, $\{4,\dots,7\}$, $\{8,\dots,13\}$, $\{14,\dots,31\}$, $\{32,\dots,63\}$, $\{64,\dots,127\}$ and $\{128,\dots,255\}$, and compares each bin with the held-out curve, interpolated linearly in $m$ and averaged over the same sizes.

\paragraph{Document recall.} Four passages of 512 tokens and four of 4,096 tokens are cut from the text at random positions. A query is a span with a uniformly random start: an 8-token cue followed by 24 tokens, whose code length is the query's, and each passage has a pool of 256 spans. The cache reads the end-of-text token, the last 960 tokens of the passage and then the span, and the fine-tune reads the end-of-text token and the span. Fine-tuning uses Adam at learning rate $10^{-4}$ on pieces of at most 256 tokens, cut at a random offset each epoch so that they cover the passage once, two pieces per step, and the pool is scored after 1, 4 and 16 epochs.

\paragraph{Anytime-valid test.} A query is drawn uniformly from the task's pool, so the per-query code lengths of the pool under both memories determine every query stream and the exact winner. The channel under test is mixed with $\varepsilon=0.01$ of the uniform law, so $c=L\log_2(|V|/\varepsilon)$: 1,425 bits for passages ($L=64$) and 534 bits for spans ($L=24$). The bet is $\lambda_j=\min\{0.75,\max\{0,\hat m_{j-1}/(\hat v_{j-1}+\hat m^2_{j-1})\}\}$, with $\hat m$ and $\hat v$ the running mean and variance of the past increments divided by $c$; the mean is shrunk toward zero by one pseudo-observation, and a pseudo-variance of 0.01 is added. Both directions run on each stream at level $0.025$, and the first to reach 40 makes the call; each comparison is summarised over 2,000 streams of at most 1,000 queries. The null error is the fraction of 10,000 streams on which the same procedure, run on the increments centred to mean zero over the pool, with $c$ widened by the absolute mean, reaches 40 within 1,000 queries.

\paragraph{Storage budget.} The four keys are fixed phrases of three tokens and the eight candidates are further tokens, all drawn with a fixed seed from the single tokens that consist of a space and at least four lowercase letters. A table assigns a candidate to each key, and its context is the end-of-text token followed by ``key word,'' for the four keys. An output is GPT-2's next token after a key, read as one of the eight candidates or as other. Reading the outputs this coarsely can only lower the information and the divergences, so the inequality of Theorem~\ref{thm:storage} holds for the coarse outputs, and the budget computed from them bounds the full outputs as well. $I(\hat Y_{1:4};D)$ is exact over the $9^4$ joint outputs and all 4,096 tables, and Equation~\eqref{eq:Ihat} is averaged over 20 draws of $M$ tables (5 for $M=512$). A replacement with parameters $(k,b)$, $0\le k\le4$ and $1\le b\le3$, is the prefix that lists the first $k$ keys with words from the first $2^b$ candidates, $2^{kb}$ states in all; each table takes the state with the smallest total divergence from its cache as GPT-2 reads it, and $H(S)$ is the entropy of the chosen state when the table is uniform. The Bayes decoder of a state returns the average of the cache's output laws over the tables assigned to it.

\subsection{Tables and airfoil pressure}\label{app:domains}
Both comparisons fix their settings before any query is scored. The level is $\alpha=0.05$ per comparison, run as two tests at level $\alpha/2$ of which the first to reach $2/\alpha$ makes the call. Both predictors are mixed with $\varepsilon=0.01$ of the uniform law over their outcomes, so the test compares the two mixed predictors, to which Theorem~\ref{thm:eprocess} applies directly. The bet on increment $j$ is $\lambda_j=\min\{\max\{m_j/(v_j+m_j^2),0\},0.75\}$, from the running mean $m_j$ and variance $v_j$ of the past normalised increments $\Delta/c$, shrunk by one pseudo-observation of variance 0.01. Each query's length is bounded in advance, one label of two classes for a table and 64 values of 482 outcomes each for an airfoil, which gives $c=\log_2(2/\varepsilon)=7.64$ and $c=64\log_2(482/\varepsilon)=995.6$ bits. Each cohort is scored once in a designated random order and replayed in 200 further random orders without repetition; the orders add no data.

\paragraph{Tables.} OpenML phoneme (1489) and spambase (44), with duplicate feature vectors removed, give 4,096 and 4,079 held-out rows after 128 training rows are drawn in each of three seeds. TabPFN-v2, the public classification checkpoint with two estimators, reads the 128 rows in its context; XGBoost, with 100 trees of depth 3, trains on them.

\paragraph{Airfoil pressure.} The AirfRANS derivative holds 1,000 simulations. Each design has 18 inputs, the inlet speed, the angle of attack and 16 samples of the surface geometry, and 64 pressure coefficients at 32 chord positions on each surface. One case represents each geometry group. The 254 query designs come from groups absent from every earlier training, calibration and test set, and the 128 training and 128 calibration designs from the groups used before. The in-context learner averages the support designs' fields with weights from a Gaussian kernel on standardised inputs, with the bandwidth chosen from six values by leave-one-out prediction on the support set; the surrogate is a two-layer MLP of width 64 with early stopping. Each gets a Gaussian spread per pressure value from its squared errors on the calibration designs, floored at 0.05 times their median, and both laws are discretised into 480 bins of width 0.05 over $[-12,12]$ standardised units and two open tails. The learning curves rerun this pipeline with nested sets of 4 to 512 designs, the first 128 being those of the seed's run, keep that run's calibration designs, queries and standardisation, and reproduce its 128-design predictions exactly.

\paragraph{Simulated streams.} Each case draws 10,000 independent streams of up to 4,096 increments with $c=7.64$: zero for the exact tie; uniform on $[-c,c]$ for the symmetric null; $0.1$ with probability $c/(c+0.1)$ and $-c$ otherwise for the skewed null; uniform on $[-c,c]$, scaled by 0.02 over the first 256 queries, for the null of changing variance; and the advantage plus independent $\pm2$ bits for the rest.

The TabPFN-v2 checkpoint is distributed under the Prior Labs licence, the OpenML datasets under their own terms, and the AirfRANS database and its derivative under the Open Database Licence 1.0.

\subsection{Learning curves in context}\label{app:counts}
Each sweep fixes its queries, items and orders before any subset is scored. Every subset that an order reaches is run through the model, which rebuilds its memory from that subset alone, with no fitted decoder and no parameter update; a subset identical to one of an earlier run reuses that run's saved predictions.

\paragraph{Occupancy.} Each shape of the eight constructed chair pairs of Appendix~\ref{app:geometry} has 300 fixed anchor points and nine optional patches of 300 points, the first of which observes the feature. The orders are four random orders drawn with a fixed seed, the order with the feature patch first and the order with it last, which reach 48 distinct subsets per pair. The code length is that of the occupancy at the 1,024 feature sites of each shape, averaged over the two shapes of the pair.

\paragraph{Stereo.} Reference view 0 of each of the five DTU scenes of Appendix~\ref{app:mvs} is reconstructed from every nonempty subset of its four source views. The code length is that of the depth hypothesis nearest the oracle depth among the network's 192, at the pixels where at least one source view observes the surface point and at least one does not, by visibility labels computed independently of the network. The observing views first and the non-observing views first are ordered for each pixel, with ties in view order, and the random curve is the mean over all subsets of each size.

\paragraph{Kriging.} Each of eight DrivAerNet query cars has six support cars from its family, and the kriging learner of Appendix~\ref{app:fields} needs two of them for its leave-one-out selection, so its curves start at two. The orders are four random orders and the order of the gains $I_G$ that the learner's own full-context predictions assign each car, computed before the sweep. The code length is the Gaussian log-density of the pressure in fixed $C_p$ units, weighted by area over 512 sites.

\subsection{Numerical conventions}\label{app:numerics}
The scorer for the occupancy, depth and pressure predictors computes categorical and Bernoulli log probabilities in a stable log domain. A zero ablated mass with positive full-context mass gives an infinite gain, which is kept. Gaussian variances are strictly positive. Numerical tests allow a declared tolerance for round-off and check that claim-level bits never exceed the gain. For a categorical law on common physical bins $d_1,\dots,d_m$,
\begin{equation}\label{eq:categorical}
I_G=\sum_{j=1}^{m}p_j\log_2(p_j/q_j),\qquad p(\cA)=\sum_{j:\,d_j\in \cA}p_j.
\end{equation}
For scalar Gaussian laws $\mathcal{N}(\mu_1,v_1)$ and $\mathcal{N}(\mu_0,v_0)$,
\begin{equation}\label{eq:gaussian}
I_G=\frac{1}{2\ln 2}\left[\ln\frac{v_0}{v_1}+\frac{v_1+(\mu_1-\mu_0)^2}{v_0}-1\right],
\end{equation}
and the probability of a fixed interval $[a-\epsilon,a+\epsilon]$ is $\Phi\big(\frac{a+\epsilon-\mu}{\sqrt v}\big)-\Phi\big(\frac{a-\epsilon-\mu}{\sqrt v}\big)$, computed from whichever tail avoids cancellation. For occupancy the local law is already Bernoulli, so the gain and the claim-level bits agree.

\section{Extended related work}\label{app:related}
\paragraph{Code length and prequential learning curves.}
Minimum description length identifies learning with compression \citep{rissanen1984universal,grunwald2007mdl}, and the prequential approach scores a forecaster by the code length of its sequential predictions \citep{dawid1984prequential}. \citet{blier2018description} showed that deep networks compress their training data far better under prequential codes than under variational ones, and \citet{deletang2024compression} showed that large language models are strong general-purpose compressors. The present account carries this view from the weights to the context: the value of a context is the code length it saves, the learning curves of the context memory and of the fine-tune are both prequential code lengths of the context (Proposition~\ref{prop:preq}), and a transformer's curve in context costs one teacher-forced pass.

\paragraph{Usable information.}
\citet{xu2020usable} define the information a computationally bounded observer can use, $\mathcal V$-information, as the reduction in the best log loss a predictive family achieves, and \citet{ethayarajh2022vusable} measure its pointwise form on single examples. The gain $\Gamma$ is pointwise usable information with the family reduced to one frozen model and its memory, and Theorem~\ref{thm:law} ties it exactly to Shannon information: its expectation is the information the memory stores, plus the prior repair, minus the decoding loss of the model's readout. Pointwise, one context can save more bits than the entropy of the query (Appendix~\ref{sec:feature}); in expectation, the information term cannot.

\paragraph{In-context learning as inference and as optimisation.}
In-context learning has been read as implicit Bayesian inference \citep{xie2022explanation}, as a learned algorithm for simple function classes \citep{garg2022can}, as learning algorithms such as gradient descent and ridge regression run by attention \citep{vonoswald2023,akyurek2023learning}, and as implicit fine-tuning \citep{dai2023gpt}; for one layer of linear self-attention, one step of gradient descent is the optimal in-context learner \citep{mahankali2024}. From the side of training, a linear layer trained by gradient descent answers a test input by attending over its training patterns \citep{irie2022dual}. Proposition~\ref{prop:cache} gives the exact form for softmax attention: with the context/prefix split of \citet{he2022unified} and the divided-difference identity of \citet{huang2026rope}, the cache is a weight write of rank at most $\min(|D|,d_h)$ whose per-example step sizes are recomputed for every query, and the law prices that write and a fine-tune in the same bits.

\paragraph{Fine-tuning as a channel.}
Under gradient flow, the fine-tune's gain on a query is a time integral of the empirical neural tangent kernel \citep{jacot2018ntk} contracted with the residuals of the query and the context (Proposition~\ref{prop:tune}), and \citet{malladi2023kernel} show that prompt-based fine-tuning of language models often follows such kernel dynamics, with the kernel extended to Adam. The transfer integral normalises the same kernel by the context's own and turns it into a budget: every bit removed from the context's code length moves $r_t$ bits to the query (Proposition~\ref{prop:transfer}). Prefix tuning and low-rank adaptation write a small state into the network \citep{li2021prefix,hu2022lora}, and \citet{he2022unified} place them in one family by writing prefix tuning as a gated mixture of attention over the prefix and over the input; the storage budget (Theorem~\ref{thm:storage}) bounds how much of the cache any such state can reproduce.

\paragraph{Context distillation and the choice between channels.}
Context distillation fine-tunes a model to reproduce, without the context, the predictions it makes with it \citep{askell2021general,snell2022learning}. It minimises a divergence of the form on the left of Theorem~\ref{thm:storage}, and the theorem sets the floor of that divergence, averaged over contexts, for a state that stores a given number of bits about the context. \citet{lampinen2025generalization} compare in-context learning and fine-tuning on knowledge held out of pretraining, find that with matched data in-context learning generalises several kinds of inference more flexibly, and improve fine-tuning by adding in-context inferences to its data; \citet{mosbach2023fewshot} compare the two at matched model size and number of examples. Proposition~\ref{prop:choose} settles the choice from one measurement whenever a certificate fires, Theorem~\ref{thm:eprocess} settles it from held-out queries with anytime-valid error control, and the transfer integral gives the route by which a fine-tune that compresses its data moves a query.

\paragraph{Anytime-valid comparison of forecasters.}
Ville's inequality bounds the chance that a nonnegative supermartingale ever crosses $1/\alpha$ \citep{ville1939}, and testing by betting builds on it tests and confidence sequences that stay valid at any stopping time \citep{ramdas2023gametheoretic,waudbysmith2024betting}. \citet{henzi2022valid} construct e-values for score differences between competing probability forecasts, and \citet{choe2024comparing} construct anytime-valid confidence sequences and e-processes for the average score difference between two sequential forecasters. Theorem~\ref{thm:eprocess} is such a comparison, with the context memory and the fine-tune as the two forecasters and code length as the score.

\section{Reproducibility}\label{app:repro}
The supplementary material holds the code and saved results behind the experiments, and its README maps each script to the section, figure and table it reproduces.

\paragraph{The package.} The package \texttt{shippable} in the supplementary material implements the evaluator, the certificates, the anytime-valid test and the adjudicator for any model that returns a probability law. Its saved laws and stopping traces for the comparisons of Section~\ref{sec:domains} replay without model weights, and \texttt{scripts/domains/airfrans\_curves.py} recomputes the learning curves of Figure~\ref{fig:domains}a from the included AirfRANS derivative. The learning curves in context of Figure~\ref{fig:counts} are recomputed by \texttt{scripts/counts/count\_curves.py} from the network outputs saved for every subset of context items.

\paragraph{Scripts.}
{\sloppy \texttt{scripts/closed\_forms.py} evaluates Equations~\eqref{eq:bayes}--\eqref{eq:gd}, the ordering and the crossover ratios, and \texttt{scripts/simulate\_and\_plot.py} runs the simulations and draws Figure~\ref{fig:closed}. In \texttt{scripts/decision/}, \texttt{train\_icl.py} trains the in-context learner, \texttt{decision\_test.py} runs every channel and computes the exact information terms, \texttt{decision\_extend.py} and \texttt{decision\_extend2.py} widen the recipe grids, \texttt{analyze\_decision.py} issues the certificates and \texttt{plot\_decision.py} draws Figure~\ref{fig:decisionfull}; the trained weights and every per-task result are in \texttt{scripts/decision/results/}. In \texttt{scripts/gpt2/}, \texttt{setup\_gpt2.py} downloads and checks the GPT-2 weights, the vocabulary and the text; \texttt{run\_shakes.py}, \texttt{run\_docs.py} and \texttt{run\_storage.py} run the three experiments of Section~\ref{sec:gpt2}; \texttt{analyze\_shakes.py}, \texttt{analyze\_eprocess.py} and \texttt{analyze\_docs.py} compute the reported numbers; and \texttt{plot\_gpt2.py} draws Figure~\ref{fig:gpt2}. Every per-stream, per-passage and per-table result is in \texttt{scripts/gpt2/results/}. Repeating the model executions requires the original datasets and pretrained weights.\par}

\end{document}